%% file: main.tex
\documentclass[10pt,twocolumn,letterpaper]{article}

\usepackage[pagenumbers]{cvpr}              

\input{preamble}

\definecolor{cvprblue}{rgb}{0.21,0.49,0.74}
\usepackage[pagebackref,breaklinks,colorlinks,allcolors=cvprblue]{hyperref}

\def\confName{CVPR}
\def\confYear{2026}

\title{Fast Spatial Memory with Elastic Test-Time Training} 

\input{authors}

\begin{document}

\twocolumn[{%
\renewcommand\twocolumn[1][]{#1}%
\maketitle
\input{floatings/fig_teaser}
}]

\def\thefootnote{$^{*}$}
\footnotetext{Authors contribute equally to this work.}
\renewcommand{\thefootnote}{\arabic{footnote}}
\setcounter{footnote}{0}


\input{sec/0_abstract}


\input{sec/1_intro}
\input{sec/3_prelim}

\input{sec/4_method}

\input{sec/5_experiment}

\input{sec/2_related}
\input{sec/6_conclusion}
{
    \small
    \bibliographystyle{ieeenat_fullname}
    \bibliography{main}
}

\clearpage
\appendix
\input{sec/a1_method}
\input{sec/a2_results}

\end{document}

%% file: preamble.tex
\usepackage[dvipsnames,table]{xcolor}

\usepackage{graphicx}
\usepackage{amsthm,amsmath,amssymb}

\usepackage{tikz}
\usepackage{algorithm}
\usepackage{comment}
\usepackage{color}
\usepackage{bbm, dsfont}

\usepackage{cutwin}
\usepackage{arydshln}
\usepackage{wrapfig}
\usepackage{bm}
\usepackage{eqparbox}
\usepackage{makecell}
\usepackage{threeparttable}

\usepackage{booktabs}
\usepackage[utf8]{inputenc} 
\usepackage[T1]{fontenc}    
\usepackage{url}            
\usepackage{amsfonts}       
\usepackage{nicefrac}       
\usepackage{microtype}      

\usepackage{times}
\usepackage{bbding}
\usepackage{multirow}
\usepackage{tabularx}
\usepackage{caption}
\usepackage{subcaption}
\usepackage{pifont}

\usepackage{newtxtext}
\usepackage{xspace}
\usepackage{natbib}
\usepackage{appendix}
\usepackage{array,etoolbox}
\usepackage{subfloat}
\usepackage{multirow}
\usepackage{listings}
\lstdefinestyle{mocov3}{
  backgroundcolor=\color{white},
  basicstyle=\fontsize{8pt}{8pt}\ttfamily\selectfont,
  columns=fullflexible,
  breaklines=true,
  captionpos=b,
  commentstyle=\fontsize{8pt}{8pt}\color[rgb]{0.25,0.5,0.5},
  keywordstyle=\fontsize{8pt}{8pt}\color[rgb]{0.85,0.18,0.50},
}
\renewcommand{\arraystretch}{1.5}

\newcommand{\cmark}{\checkmark}%
\newcommand{\xmark}{\ding{55}}%
\newcommand{\model}{\textsc{FSM}\xspace}

\newcommand{\boldstart}[1]{\noindent\textbf{#1}\ }
\newcommand{\boldstartspace}[1]{\medskip\noindent\textbf{#1}}

\usepackage{wrapfig}
\usepackage[normalem]{ulem}

%% file: authors.tex
\newcommand{\ibm}{$^{1}$}
\newcommand{\umich}{$^{2}$}
\newcommand{\umass}{$^{3}$}
\newcommand{\colead}{$^{*}$}
\newcommand{\andaffto}{$^,$}

\author{%
    Ziqiao Ma\ibm\andaffto\umich\colead \hspace{2pt} 
    Xueyang Yu\umass\colead \hspace{2pt}
    Haoyu Zhen\umass\hspace{2pt}
    Yuncong Yang\umass\hspace{2pt} 
    Joyce Chai\umich\hspace{2pt}
    Chuang Gan\ibm\andaffto\umass \\
    \ibm MIT-IBM Watson AI Lab \quad
    \umich University of Michigan \quad
    \umass University of Massachusetts Amherst \\
    \texttt{\url{https://fast-spatial-memory.github.io/}}
}

%% file: floatings/fig_teaser.tex
\begin{center}
    \includegraphics[width=\linewidth]{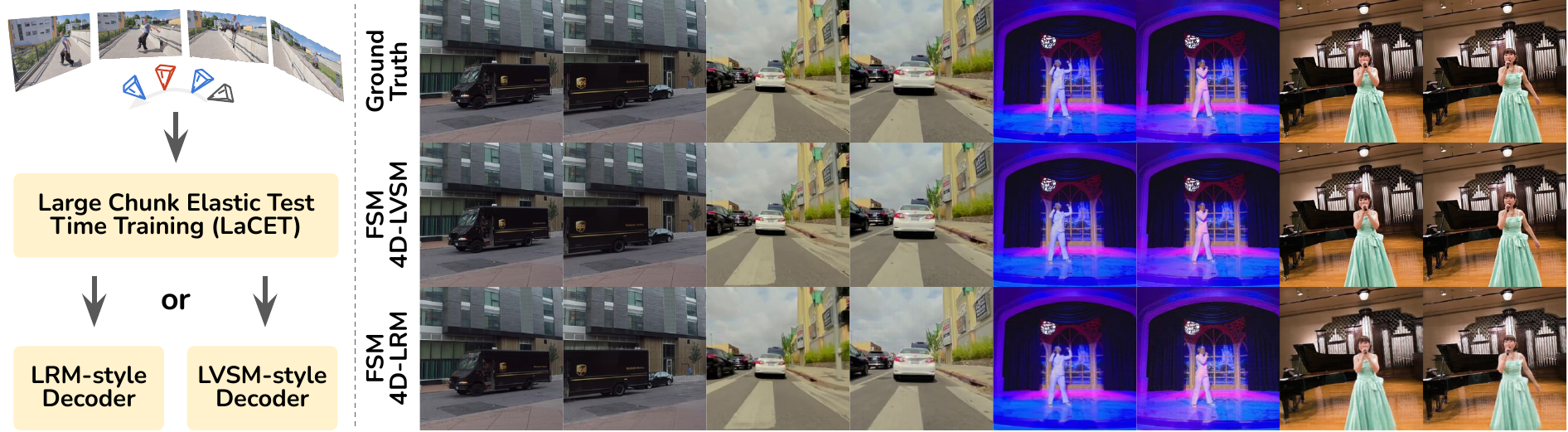}
    \vspace{-15pt}
    \captionof{figure}{Fast Spatial Memory (\model) is an efficient, scalable 4D reconstruction model that learns spatiotemporal representations from long sequences to render novel views at novel times. The model is powered by Large Chunk Elastic Test-Time Training (LaCET) blocks and is compatible with a range of rendering decoders, including LRM-style and LVSM-style decoders. \vspace{5pt}}
    \label{fig:teaser}
\end{center}

%% file: sec/0_abstract.tex
\begin{abstract}

Large Chunk Test-Time Training (LaCT) has shown strong performance on long-context 3D reconstruction, but its fully plastic inference-time updates remain vulnerable to catastrophic forgetting and overfitting. 
As a result, LaCT is typically instantiated with a single large chunk spanning the full input sequence, falling short of the broader goal of handling arbitrarily long sequences in a single pass.
We propose Elastic Test-Time Training inspired by elastic weight consolidation, that stabilizes LaCT fast-weight updates with a Fisher-weighted elastic prior around a maintained anchor state. 
The anchor evolves as an exponential moving average of past fast weights to balance stability and plasticity.
Based on this updated architecture, we introduce Fast Spatial Memory (\model), an efficient and scalable model for 4D reconstruction that learns spatiotemporal representations from long observation sequences and renders novel view-time combinations.
We pre-trained \model on large-scale curated 3D/4D data to capture the dynamics and semantics of complex spatial environments. 
Extensive experiments show that \model supports fast adaptation over long sequences and delivers high-quality 3D/4D reconstruction with smaller chunks and mitigating the camera-interpolation shortcut.
Overall, we hope to advance LaCT beyond the bounded single-chunk setting toward robust multi-chunk adaptation, a necessary step for generalization to genuinely longer sequences, while substantially alleviating the activation-memory bottleneck.

\end{abstract}

%% file: sec/1_intro.tex
\vspace{-15pt}
\section{Introduction}
\label{sec:intro}

Building a spatial memory would require learning to compress visual observations across viewpoints and time into a unified 4D representation that preserves both spatial structure and temporal dynamics. 
This capability would advance applications in 4D asset generation~\cite{xie2024sv4d,ren2024l4gm} for video games, film production, and AR/VR, as well as world modeling~\cite{kerr2024robot,zhen2025learning} for embodied AI and robotics.
Especially, reconstructing dynamic scenes from temporally extended and dynamically sampled observations (e.g., long videos captured by moving cameras) remains a central challenge.

Recent advances in Large Reconstruction Models (LRMs)~\cite{hong2024lrm,zhang2024gslrm} and Large View Synthesis Models (LVSM)~\cite{jin2025lvsm,kim2026scaling} offer promising rendering-based alternatives to efficient and high-quality 3D/4D reconstruction. 
Typically built on Transformer-based sequence modeling, these methods achieve strong reconstruction performance by learning powerful priors over structure and appearance from large-scale multi-view data.
Despite these advances, these models remain constrained by the amount of activation memory available for a single forward pass, leaving long-context modeling largely unresolved.
This is particularly the case in 4D domain, where videos that are temporally extended yet spatially sparsely observed, and their reconstruction quality degrades sharply beyond the training context length, indicating limited temporal scalability~\cite{ma20254d}. 
While several 3D reconstruction works have explored hybrid sequence models that combine linear-time state-based mixers with full attention~\cite{ziwen2025long,ziwen2025longplus}, the central question for practical 4D modeling remains open:
\textit{How can we design a simple, scalable, and efficient spatial memory architecture that learns scene-level spatiotemporal representations from long sequences?}

Test-Time Training (TTT)~\cite{schlag2021linear,sun2025learning} has shown promise in addressing the long-context issue in geometric reconstruction and view synthesis~\cite{chen2026ttt3r,zhang2026loger,wang2026tttlrm}.
Especially, Large Chunk Test-Time Training (LaCT)~\cite{zhang2025test} enables in-forward, chunk-wise fast-weight adaptation that lets a transformer recalibrate its internal representations during inference, efficiently updating small parameters from key-value statistics without backpropagation to achieve self-refining, test-time adaptation.
Yet, these techniques do not directly generalize to the 4D regime, where scene dynamics evolve across space and time during inference, since the fully plastic nature of continuous LaCT updates leads to uncontrolled fast-weight drift, leading to overfitting in training and unstable updates at test time.
This is analogous to catastrophic forgetting at inference time.
To address this issue, we introduce Elastic Test-Time Training that executes an additional \textit{consolidate} operation after LaCT \textit{update}, inspired by the Elastic Weight Consolidation (EWC)~\cite{kirkpatrick2017overcoming} in continual learning.
Each fast-weight module keeps a reference set of anchor parameters (the values before adaptation) and continuously estimates their importance through an online Fisher-style statistic.
During inference, important parameters are softly pulled back toward their anchors, while less critical ones remain free to adjust.
This elastic behavior acts as an adaptive spring: it constrains unstable drift without sacrificing responsiveness to new lighting, pose, or scene conditions, transforming the base transformer into a fast, self-refining yet elastic 4D learner, one that keeps adapting to the stream while remembering where it came from.
We refer to this new architecture as Large Chunk Elastic Test-Time Training (LaCET).

We scale LaCET up to pretrain a Fast Spatial Memory (\model) on a curated set of 3D/4D datasets with posed images captured over time and from different cameras. 
We primarily evaluated \model on the novel view synthesis (NVS) and demonstrated its competitive performance on a variety of benchmarks and the scalability of LaCET.
The model scales effectively with more data and larger model size and generalizes well to novel scenes.
With careful ablation studies, we show that LaCET can effectively mitigate the overfitting and undesirable inference time behaviors of LaCT, e.g., camera interpolation.
To our knowledge, \model is the first large-scale 4D reconstruction model design that supports input from long sequences of views and arbitrary timestamps and renders arbitrary novel view-time combinations. 
Overall, we hope to advance LaCT beyond the bounded single-chunk setting toward robust multi-chunk adaptation, a necessary step for generalization to genuinely longer sequences, while substantially alleviating the activation-memory bottleneck.

%% file: sec/3_prelim.tex
\section{Algorithmic Preliminaries}
\label{sec:prelim}

\input{floatings/fig_model}
\subsection{Fast Weights and Test-Time Training}

Test-Time Training (TTT)~\cite{sun2025learning} introduces fast weights~\cite{schlag2021linear} with rapidly adaptable parameters, which get updated at both training and inference time.
This is in sharp contrast to slow weights (conventional model parameters) that remain fixed at inference time.
In the context of attention, we consider a sequence of $N$ tokens $\mathbf{x} = [x_1, x_2, \dots, x_N]$, where each token $x_i$ is projected into key $k_i$, query $q_i$, and value $v_i$ vectors. 
Formally, TTT defines a function $f_{\boldsymbol{\theta}}(\cdot)$ parameterized by the fast weights $\boldsymbol{\theta}$, and it involves an \textit{update} and an \textit{apply} operation.
The (per-token) \textit{update} operation defines:
\begin{equation}
\boldsymbol{\theta}' = \boldsymbol{\theta} - \eta \, \nabla_{\boldsymbol{\theta}} \mathcal{L}\big(f_{\boldsymbol{\theta}}(k_i), v_i\big),
\label{eq:ttt_update}
\end{equation}
where $\eta$ represents the learning rate and $\mathcal{L}(\cdot,\cdot)$ denotes a loss between the transformed key $f_{\boldsymbol{\theta}}(k_i)$ and its corresponding value $v_i$, encouraging the network to learn key-value associations.
Intuitively, this objective trains the model to compress the ever-growing KV cache (whose memory cost scales linearly with context length) into a fixed-size neural memory, preserving critical key-value associations within a bounded memory budget.
The \textit{apply} operation defines:
\begin{equation}
z_i = f_{\boldsymbol{\theta}'}(q_i),
\label{eq:ttt_apply}
\end{equation}
where the updated fast weights $\boldsymbol{\theta}'$ are used to compute the output vector $z_i$ given the query $q_i$.
The per-token TTT layer iteratively performs the \textit{update} and \textit{apply} operations on each token $x_i$ in sequence.

\subsection{Test-Time Training Done Right}
Na\"ive TTT methods often struggle to scale to long contexts, largely due to the low hardware efficiency of their TTT layers, which operate on extremely small mini-batches.
To address this, \cite{zhang2025test} proposed Large-Chunk Test-Time Training (LaCT), a chunk-wise formulation that improves scalability and throughput.
The \textit{apply} operation $o_i = f_{\boldsymbol{\theta}}(q_i)$ follows Eq.~\eqref{eq:ttt_apply}, where all query vectors ${q_i}$ within a chunk share the same fast weight.
Unlike the per-token update in Eq.~\eqref{eq:ttt_update}, LaCT aggregates the loss over all keys $k_i$ and values $v_i$ in a chunk and computes a single surrogate \textit{update} for chunk $c$:
\begin{equation}
\boldsymbol{\theta}_{c+1}
\;=\;
\boldsymbol{\theta}_c \;
\underbrace{
    -\;
    \left.
    \nabla_{\boldsymbol{\theta}}
    \sum_{i=1}^{b}
    \eta_i(x_i)\,
    \mathcal{L}\big(f_{\boldsymbol{\theta}}(k_i), v_i\big)
    \right|_{\boldsymbol{\theta}=\boldsymbol{\theta}_c}
}_{\text{per-chunk surrogate pseudo-gradient}}.
\label{eq:lact_update}
\end{equation}
Here, $b$ denotes the chunk size and $\eta_i$ is the (learnable) per-token learning rate.
Intuitively, this objective strengthens the association between each key and its corresponding value by updating the fast weights so that $f_{\boldsymbol{\theta}}(k_i)$ becomes more consistent with $v_i$ under the training loss.
In practice, LaCT regularizes the updated fast weights using L2 weight normalization~\cite{salimans2016weight} along the input dimension and optionally applies the Muon-style Newton-Schulz iteration~\cite{jordan2024muon,liu2025muon}, without weight decay.
Because each chunk aggregates thousands of tokens, updates occur infrequently, enabling richer update-rule designs while amortizing computational cost.

\subsection{Test-Time Training Done Better}

While LaCT significantly improves the scalability of TTT by amortizing adaptation across large chunks, its updates remain fully plastic, as the fast weights in each chunk drift freely in parameter space at inference time.
In the novel view synthesis task, LaCT works the best with one single chunk.
In long and dynamic 4D scenes, where illumination, pose, or motion continuously evolve during inference, such unconstrained plasticity can cause cumulative instability, leading to temporal ghosting artifacts.
To address this, we propose Elastic Test-Time Training, which enhances the LaCT update operator with an Elastic Weight Consolidation (EWC)~\cite{kirkpatrick2017overcoming} regularizer, introducing a soft stability prior over fast-weight dynamics.
We refer to our algorithm as Large-Chunk Elastic Test-Time Training (LaCET, to distinguish from LaCT), combining its scalability, efficiency, and elastic stability for robust long sequence modeling.

\boldstartspace{Elastic Weight Consolidation.}
\citet{kirkpatrick2017overcoming} introduces a quadratic penalty that discourages important parameters from drifting too far from a reference set of anchor weights, originally designed for a classic continual learning setting where a model learn a new task $\mathcal{T}_B$ without forgetting a previously learned task $\mathcal{T}_A$.
All knowledge about $\mathcal{T}_A$ is captured in the posterior distribution $p(\boldsymbol{\theta}\,|\,\mathcal{D}_A)$.
Since this posterior is intractable for large neural networks, EWC approximates it using a Gaussian centered at the previously optimized parameters
$\boldsymbol{\theta}_A^{\star}$ with a diagonal precision given by the Fisher Information Matrix $F$, i.e., $p(\boldsymbol{\theta}\,|\,\mathcal{D}_A) \approx \mathcal{N}\!\big(\boldsymbol{\theta}_A^{\star}, F^{-1}\big).$
The Fisher Information has three desirable properties:
(i) it corresponds to the local curvature of the loss near $\boldsymbol{\theta}_A^{\star}$,
(ii) it can be estimated from first-order gradients alone,
and (iii) it is guaranteed to be positive semi-definite.
The overall objective when learning $\mathcal{T}_B$ becomes
a combination of the new-task loss and a quadratic penalty at
$\boldsymbol{\theta}_A^{\star}$:
\begin{equation}
\mathcal{L}(\boldsymbol{\theta})
=
\mathcal{L}_B(\boldsymbol{\theta})
+ 
\sum_i
\frac{\lambda}{2}\,
F_i\,
\big(\boldsymbol{\theta}_i - \boldsymbol{\theta}_{A,i}^{\star}\big)^2,
\label{eq:ewc_original_ab}
\end{equation}
where $\mathcal{L}_B(\boldsymbol{\theta})$ is the loss for the new task $\mathcal{T}_B$,
$\lambda$ controls the relative importance of retaining old knowledge,
and $i$ indexes each model parameter.
Intuitively, parameters with high Fisher values $F_i$ are crucial for $\mathcal{T}_A$
and are therefore strongly constrained to remain near $\boldsymbol{\theta}_A^{\star}$,
whereas parameters with small $F_i$ can adapt freely to $\mathcal{T}_B$.

\boldstartspace{Elastic Test-Time Training.}
In our formulation, we reinterpret this idea at the time of the test: each incoming chunk of data acts as a new task $\mathcal{T}_B$, and the fast-weight state of the previous chunk plays the role of $\boldsymbol{\theta}_A^{\star}$.
The Fisher-weighted penalty in Eq.~\eqref{eq:ewc_original_ab} thus serves as a continuously updated elastic prior, stabilizing the model's adaptation over time (e.g., foreground dynamics) while preserving useful past information (e.g., static background).
The EWC penalty defines an elastic prior after the LaCT \textit{update} in Eq.~\eqref{eq:lact_update}, which we refer to as the \textit{consolidate} operator.
Formally, let $\boldsymbol{\theta}_c'$ denote the intermediate fast weights after the \textit{update} but before elastic consolidation in chunk $c$, and $\boldsymbol{\theta}_c^\star$ their corresponding \emph{anchor} parameters (the reference state before adaptation or at the last re-anchor).
\begin{equation}
\boldsymbol{\theta}_{c+1}
\;=\;
\boldsymbol{\theta}'_c \;
\underbrace{
    -\;
    \lambda\, F_c \odot
    \big(\boldsymbol{\theta}'_c - \boldsymbol{\theta}_c^\star\big),
}_{\text{elastic consolidation}}
\label{eq:ettt_update_combined}
\end{equation}
where $F_c$ is a per-parameter Fisher-style importance estimate, $\odot$ denotes the Hadamard (elementwise) product, and $\lambda$ is a constant controlling the strength of the elastic prior.

\boldstartspace{Importance Estimates.}
We maintain the importance matrix $F_c$ as an EMA with decay $\alpha \in [0,1)$ over chunk index $c$:
\begin{equation}
    F_{c+1} \;=\; \alpha\, F_c \;+\; (1-\alpha)\, \varphi\!\big(\mathbf{S}_c\big),
    \label{eq:fisher_multi_mode}
\end{equation}
where \(\alpha \in [0,1)\) is the decay factor. The statistic \(\mathbf{S}_c\) depends on the chosen estimator. 
Besides EWC~\cite{kirkpatrick2017overcoming}, we also consider two related alternatives motivated by memory-aware synapses (MAS)~\cite{aljundi2018memory} and synaptic intelligence (SI)~\cite{zenke2017continual}. 
Concretely,
\begin{align*}
    \mathbf{S}_c &=
    \begin{cases}
    \boldsymbol{\theta}'_c - \boldsymbol{\theta}_c, & (\textit{MAS / EWC}) \\
    (\boldsymbol{\theta}'_c - \boldsymbol{\theta}_c)\odot(\boldsymbol{\theta}'_c - \boldsymbol{\theta}_c^\star), & (\textit{SI})
    \end{cases} \\
    \varphi(\mathbf{S}_c) &=
    \begin{cases}
    \lvert \mathbf{S}_c \rvert, & (\textit{MAS / SI}) \\
    \mathbf{S}_c^{\,2}, & (\textit{EWC})
    \end{cases}
\end{align*}
with all operations applied elementwise. 
When \(\mathbf{S}_c\) has a leading batch dimension, we average over that dimension before applying Eq.~\eqref{eq:fisher_multi_mode}.
Intuitively, the MAS-like variant tracks the magnitude of the chunkwise update, the EWC-like variant emphasizes parameters that consistently receive large squared updates, and the SI-like variant additionally weights the update by its drift from the current anchor. 
In our setting, since the anchor-relative displacement is itself induced by the chunkwise update, the SI-like statistic tends to behave similarly to a rescaled squared-update estimator.

\boldstartspace{Anchor Update Policies.}
We consider different anchoring policies that control how $\boldsymbol{\theta}^\star$ is maintained:
\begin{itemize}[leftmargin=*]
    \setlength\itemsep{-0.05em}
    \item \textbf{Global:} anchors remain fixed to initialization.
    \item \textbf{Streaming:} anchors update at each chunk boundary, ensuring local temporal continuity.
    \item \textbf{Streaming-EMA:} anchors update via an exponential moving average~\cite{tarvainen2017mean}, $\boldsymbol{\theta}^\star \leftarrow \beta \boldsymbol{\theta}^\star + (1-\beta)\boldsymbol{\theta}$, forming a low-pass filter over the fast-weight trajectory.
\end{itemize}
We will show later that Streaming-EMA is the best practice for genuinely elastic memory behaviors.

%% file: floatings/fig_model.tex
\begin{figure*}[t!]
    \centering
    \includegraphics[width=1.0\linewidth]{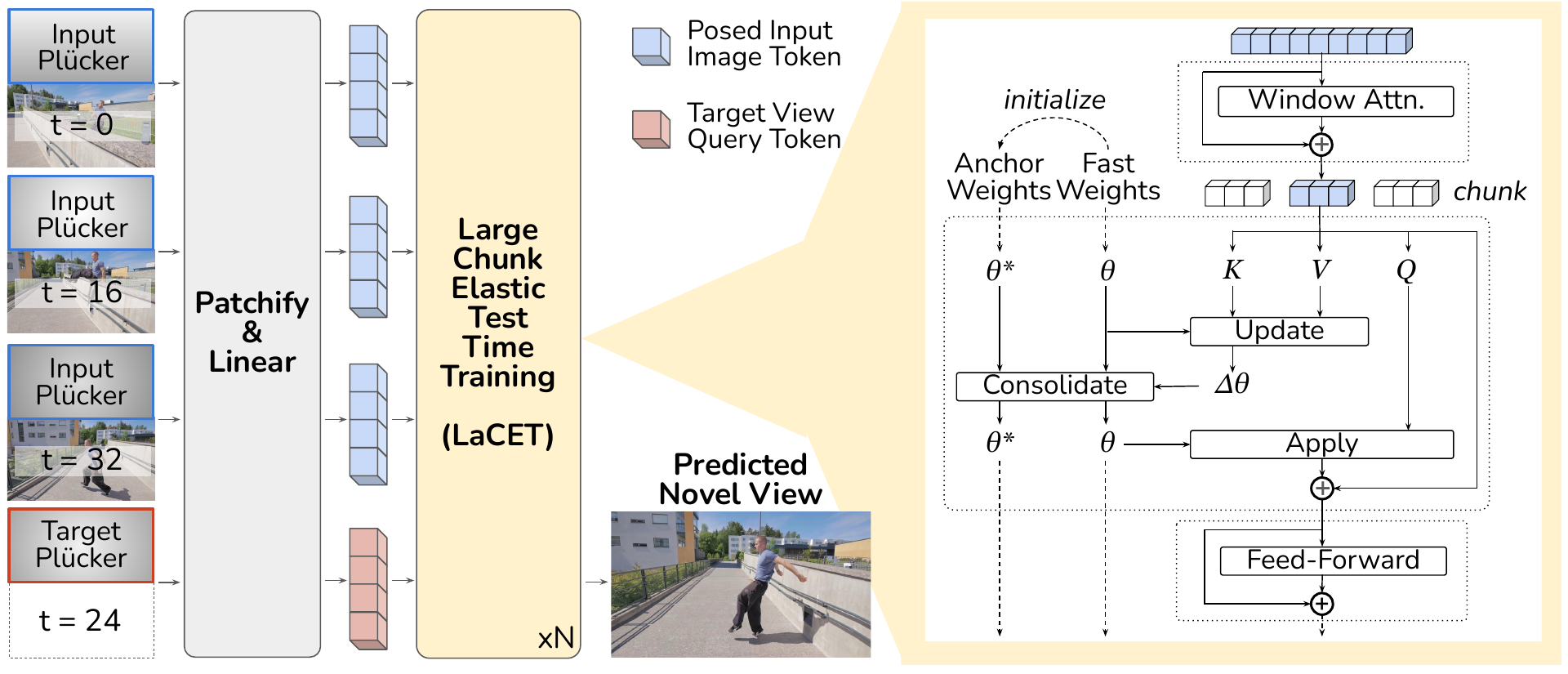}
    \vspace{-15pt}
    \caption{\textbf{(Left)} Overview of \model. The model takes a sequence of posed images captured at different times and learns to infer novel view-time combinations. Camera information is converted into Pl\"ucker ray maps as geometric augmentation for visual tokens. The model directly predict the target view with decoders. 
    \textbf{(Right)} The LaCET Block. It maintains two sets of parameters, \textit{anchor weights} and \textit{fast weights}. During adaptation, the fast weights are updated using information from the current chunk (queries, keys, and values), while the anchor weights act as a stable reference. The model tracks parameter importance online and softly restores critical weights toward their anchors to prevent drift. This stabilizes rapid updates while preserving the adaptability of TTT, addressing the plasticity issue. \vspace{-10pt}}
    \label{fig:model}
\end{figure*}

%% file: sec/4_method.tex
\section{Fast Spatial Memory (\model)}
\label{sec:method}

\input{floatings/fig_decoder}

\model adopts an end-to-end feedforward network to learn scene representations, trained using only photometric supervision.
Input images are patchified and augmented with temporal and camera information to form visual tokens, which are then processed by the sequence model. 
We consider two decoding variants: (i) direct RGB patch prediction with a lightweight linear head, in the spirit of LVSMs~\cite{jin2025lvsm,kim2026scaling}; and (ii) prediction of pixel-aligned Gaussian Splatting primitives followed by rasterization into target views, in the spirit of GS-LRMs~\cite{zhang2024gslrm,ma20254d,wang2026tttlrm}.

\subsection{Model Architecture}

\boldstartspace{Image Tokenization.}
As shown in Figure~\ref{fig:decoder}, the input consists of $V$ posed images from arbitrary view-time combinations, denoted as $\{\mathbf{I}_j \in \mathbb{R}^{H \times W \times 3}\}_{j=1}^{V}$, together with their camera intrinsics and extrinsics.
Here, $H$ and $W$ denote the image height and width, respectively.
We convert the provided camera parameters into canonical Pl\"ucker ray maps~\cite{plucker1865xvii}, represented as 
$[\mathbf{r}_d,\, \mathbf{r}_o \times \mathbf{r}_d]$, where $\mathbf{r}_d$ and $\mathbf{r}_o$ denote the ray direction and origin, respectively.
Following 4D-LRM~\cite{ma20254d}, temporal conditioning is encoded using a timestamp map $\{\mathbf{T}_j \in \mathbb{R}^{H \times W \times 1}\}_{j=1}^{V}$, which records the normalized time of each frame.
For input $j$, We concatenate the timestamp $\mathbf{T}_j$, RGB image $\mathbf{I}_j$, and Pl\"ucker ray map $\mathbf{P}_j$ along the channel dimension to form a per-view feature map $\widetilde{\mathbf{I}}_j = \mathrm{Concat}(\mathbf{I}_j,\, \mathbf{P}_j,\, \mathbf{T}_j) 
\in \mathbb{R}^{H \times W \times 10}$, which provides per-pixel spatial and temporal embeddings to distinguish both frame time and camera view.
Each $\widetilde{\mathbf{I}}_j$ is partitioned into non-overlapping patches of size $p \times p$.  
Every patch is flattened into a vector of length $10p^2$ and linearly projected to a $D$-dimensional token embedding.

\boldstartspace{LaCET Backbone.}
We adopt SwiGLU-MLP~\cite{shazeer2020glu} without bias terms as the fast weight network in Eq.~\eqref{eq:lact_update}, consisting of three parameter matrices $\boldsymbol{\theta} = \{\boldsymbol{\theta}_1,\boldsymbol{\theta}_2,\boldsymbol{\theta}_3\}$.
The network and its loss is:
\begin{align}
\mathcal{L}\big(\textcolor{blue}{f_{\boldsymbol{\theta}}(k_i)}, v_i\big)
&= -\,\textcolor{blue}{f_{\boldsymbol{\theta}}(k_i)}^{\!\top} v_i \nonumber \\
&= -\,\textcolor{blue}{
[\boldsymbol{\theta}_2
\!\left(
  \mathrm{SiLU}(\boldsymbol{\theta}_1 k_i)
  \circ
  (\boldsymbol{\theta}_3 k_i)
\right)]
}^{\!\top} v_i,
\label{eq:swiglu_loss}
\end{align}
where $\circ$ denotes elementwise multiplication.
We emphasize that only the input-view tokens are passed through the KV projections to generate gradients for the \textit{update} operation.
This design ensures that the target-view tokens do not interact with one another, allowing each novel view to be synthesized independently and efficiently.
In contrast, allowing target tokens to interact across views would correspond to a form of dynamic evaluation~\cite{krause2018dynamic} or few-shot in-context learning~\cite{von2023transformers}, which introduces additional information leakage and renders the comparison unfair.

\boldstartspace{LVSM-Style Rendering.}
In the LVSM-style variant (Figure~\ref{fig:lvsm}), the model does not rely on an explicit scene representation. 
For each target view-time query, we construct an empty image-token map whose appearance channels are set to zero, while its camera and temporal channels are populated with the target metadata. 
These query tokens are concatenated with the input tokens and processed jointly by the model. 
We then use a lightweight image-token decoder to reconstruct RGB patches from the output token embeddings. 
Concretely, each token is first passed through layer normalization, then projected linearly from the token dimension to $3p^2$. 
The resulting vector is interpreted as the flattened RGB values of the reconstructed patch.
The resulting vector is interpreted as the flattened RGB values of the reconstructed patch, followed by a sigmoid activation to bound predictions to $[0,1]$ in normalized pixel space.

\boldstartspace{(Alternatively) LRM-Style Rendering.}
Following an LRM-style rendering (Figure~\ref{fig:lrm}), we adopt an explicit 4D representation, e.g., 4DGS~\cite{yang2024realtime} similar to 4D-LRM~\cite{ma20254d}.
To adapt the sequence model for explicit GS modeling, we follow tttLRM~\cite{wang2026tttlrm} to query the fast weights for a set of virtual view planes for 4DGS and use the input views as virtual views.
We adopt pixel-aligned Gaussian rendering, leading to $V \times H \times W$ Gaussian primitives, each parameterized by $\mathbf{g}\in \mathbb{R}^{20}$.
We split it into $(\mathbf{g}_\mathrm{xyz}\in \mathbb{R}^{3}, \mathbf{g}_\mathrm{t}\in \mathbb{R}, \mathbf{g}_\mathrm{rgb}\in \mathbb{R}^{3},\mathbf{g}_\mathrm{scale,xyz}\in \mathbb{R}^{3}, \mathbf{g}_\mathrm{scale,t}\in \mathbb{R},\mathbf{g}_\mathrm{rotation,left}\in \mathbb{R}^{4}, \mathbf{g}_\mathrm{rotation,right}\in \mathbb{R}^{4},\mathbf{g}_\mathrm{opacity}\in \mathbb{R})$.
We mostly followed the parameterization of 4D-LRM except we set the permissible depth interval $\delta_\mathrm{near} = 0.01$ and $\delta_\mathrm{far} = 100$ for scene-level reconstruction.
We adopt tile-based rasterization with deferred backpropagation during rendering to reduce GPU memory consumption~\cite{zhang2022arf}.

\input{floatings/fig_ablation}
\input{floatings/tab_ablation}

\subsection{Training Objectives}
To train the model, we render $U$ target views for supervision and minimize the image reconstruction loss. 
Let $\{ \mathbf{I}^*_{i'} \mid i' = 1, 2, \ldots, U \}$ denote the ground truth views and $\{ \widehat{\mathbf{I}}^*_{i'} \}$ the corresponding rendered images. 
The photometric training loss combines $\ell_2$ (MSE) loss and LPIPS (w/ VGGNet) loss~\cite{zhang2018unreasonable}:
\begin{equation}
\mathcal{L} = \frac{1}{U} \sum_{i'=1}^{U} \left( \ell_2(\widehat{\mathbf{I}}^*_{i'}, \mathbf{I}^*_{i'}) + \mu \cdot \mathrm{LPIPS}(\widehat{\mathbf{I}}^*_{i'}, \mathbf{I}^*_{i'}) \right),
\end{equation}
where $\mu$ controls the weight of the LPIPS loss and is set to 0.5 empirically.

\input{floatings/tab_data}

\subsection{Pretraining Dataset}

A summary of the datasets used for pretraining is provided in Table~\ref{tab:dataset}, including RealEstate10K~\cite{zhou2018stereo},  DL3DV~\cite{ling2024dl3dv}, PointOdyssey~\cite{zheng2023pointodyssey}, Spring~\cite{mehl2023spring}, DynamicReplica~\cite{karaev2023dynamicstereo}, Multi-Cam Video~\cite{bai2025recammaster}, and Stereo4D~\cite{jin2025stereo4d}.
Due to the limited availability of 4D data, we retain several static datasets and assign timestamps according to the natural camera trajectory.
For other synthetic datasets, frame timestamps are randomly assigned to each view.
All datasets are rescaled to maintain a consistent metric scale across sources.
Data pre-processing details are in Appendix~\ref{app:preprocess}.

%% file: floatings/fig_decoder.tex
\begin{figure*}[!t]
\centering
    \begin{subfigure}[t]{.49\textwidth}
        \centering
        \includegraphics[width=1.0\textwidth]{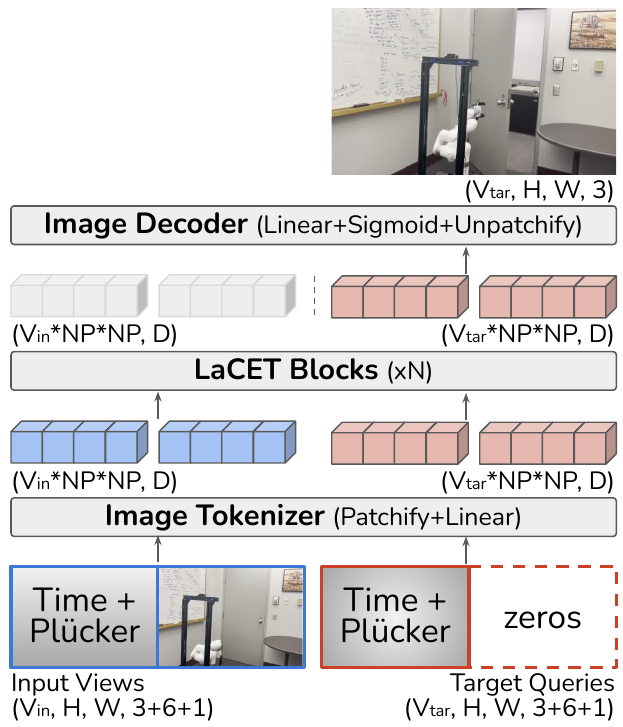}
        \vspace*{-10pt}
        \caption{\model-LVSM Overview.}
        \label{fig:lvsm}
    \end{subfigure}
    ~
    \hspace{-5pt}
    \begin{subfigure}[t]{.49\textwidth}
        \centering
        \includegraphics[width=1.0\textwidth]{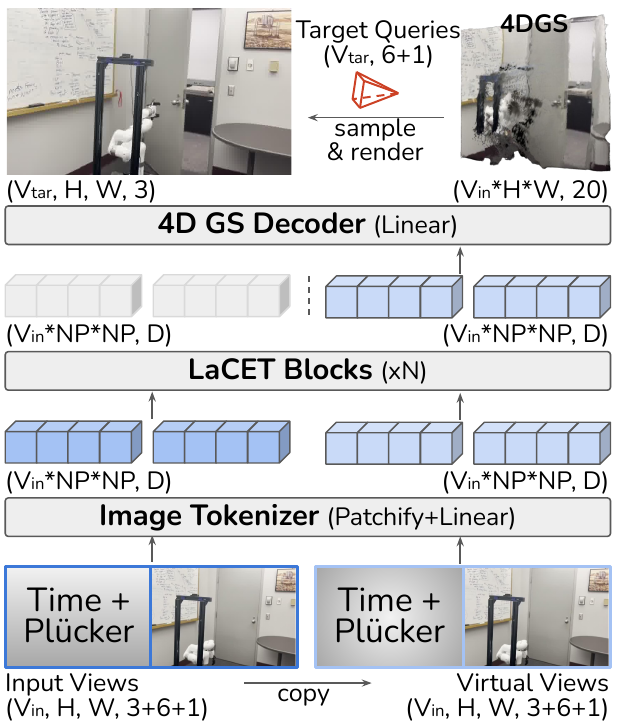}
        \vspace*{-10pt}
        \caption{\model-LRM Overview.}
        \label{fig:lrm}
    \end{subfigure}
    \vspace{-10pt}
    \caption{
    \model-LVSM and \model-LRM architectural designs. (a) LVSM-style rendering predicts target image patches directly from query tokens and does not build an explicit scene representation. (b) LRM-style rendering first predicts an explicit 4D scene representation with Gaussian primitives and then renders target views from that representation. \vspace{-10pt}
    }
    \label{fig:decoder}
\end{figure*}

%% file: floatings/fig_ablation.tex
\begin{figure*}[t!]
    \centering
    \includegraphics[width=1.0\linewidth]{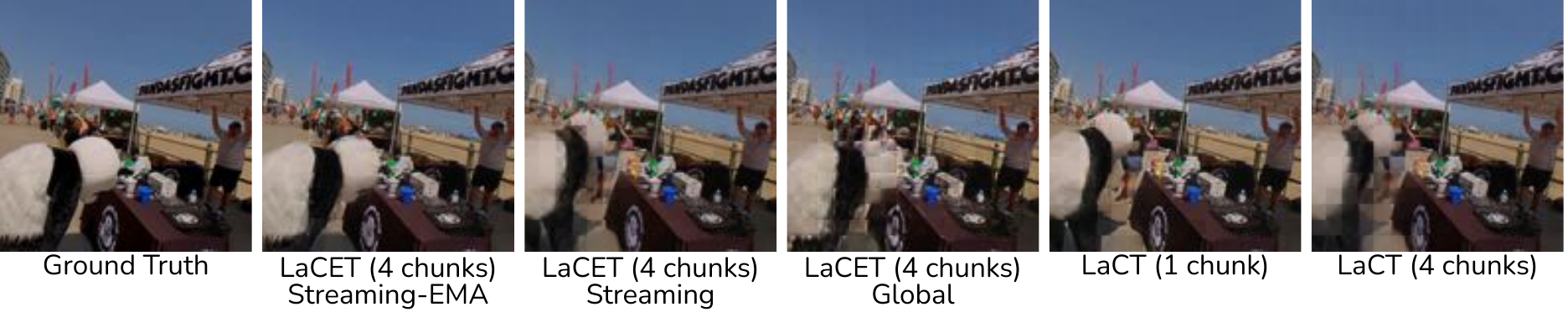}
    \vspace{-20pt}
    \caption{Qualitative illustration of the ablation studies, obtained after the same training steps (16K) with the same training and inference random seed on the same Stereo4D test set example. \vspace{-5pt}}
    \label{fig:ablation}
\end{figure*}

%% file: floatings/tab_ablation.tex
\begin{table*}[!t]
    \centering
    \scalebox{0.9}{
    \hspace{-12pt}
    \begingroup
    \renewcommand{\arraystretch}{0.9}
    \setlength{\tabcolsep}{8pt}
    \begin{threeparttable}
    \begin{tabular}{cccccccccc}
    \toprule
    \multirow{2}{*}{\textbf{EWC}} &
      \textbf{Train} &
      \textbf{Test} &
      \textbf{Test} &
      \textbf{Anchor} &
      \textbf{Fisher} &
      \textbf{Train} &
      \multicolumn{3}{c}{\textbf{Test}} \\ [-2pt]
      \cmidrule(lr){8-10}
           & 
      \textbf{\#Chunk} &
      \textbf{\#Chunk} &
      \textbf{Batch Size} &
      \textbf{Update} &
      \textbf{Estimate} & \textbf{$\ell_2$ Loss ($\times 10^3$)}$^\downarrow$ & \textbf{PSNR}$^\uparrow$ & \textbf{LPIPS}$^\downarrow$ & \textbf{SSIM}$^\uparrow$ \\ 
    \cmidrule(lr){1-6}\cmidrule(lr){7-7}\cmidrule(lr){8-10}
    \xmark & 1 & 1 & 1  & -             & -      & 1.80             & 26.021        & 0.1179          & 0.792         \\
    \xmark & 4 & 4 & 1  & -             & -      & 2.04             & 26.908        & 0.0988          & 0.814         \\
    \cmidrule(lr){1-6}\cmidrule(lr){7-7}\cmidrule(lr){8-10}
    \cmark & 4 & 4 & 1  & streaming-ema & SI    & 2.36             & \textbf{29.989}        & \textbf{0.0517}          & \textbf{0.903}        \\
    \cmark & 4 & 4 & 1  & streaming-ema & EWC    & 2.36             & 29.781        & 0.0537          & 0.897        \\
    \cmark & 4 & 4 & 1  & streaming-ema & MAS & 2.28             & 29.922        & 0.0519          & 0.899         \\
    \cmark & 4 & 4 & 1  & streaming     & MAS & 1.71             & 26.960        & 0.0966          & 0.817         \\
    \cmark & 4 & 4 & 1  & global        & MAS & 3.00             & 28.347        & 0.0653          & 0.863         \\
    \cmark & 1 & 1 & 1  & global$^*$        & MAS & 1.73             & 26.965        & 0.0960          & 0.817         \\
    \cmark & 1 & 4 & 1  & streaming-ema & MAS & 1.73             & 21.993        & 0.3429          & 0.650         \\
    \cmark & 4 & 4 & 16 & streaming-ema & MAS & 2.28             & 29.928        & 0.0519          & 0.898         \\
    \bottomrule
    \end{tabular}
    \begin{tablenotes}
    \item[*] \textit{The choice of anchor update policy makes no difference when chunk size is set to full sequence.}
    \end{tablenotes}
    \end{threeparttable}
    \endgroup}
    \vspace{-5pt}
    \caption{
        \textbf{Ablation Studies.}
        The training $\ell_2$ loss is reported from the exponential moving average (EMA) model ($\alpha = 0.1$) to ensure robustness against noise. 
        When the number of chunks is $1$, it corresponds to the original full-sequence setup in LaCT. With $4$ chunks, each chunk contains $2048$ input tokens. 
        We find that EWC effectively mitigates the overfitting issue observed in LaCT due to full plasticity.
        The \textit{streaming-ema} anchor update policy proves critical for achieving stable performance.
        \vspace{-10pt}
    }
    \label{tab:ablation}
\end{table*}

%% file: floatings/tab_data.tex
\begin{table}
    \centering
    \hspace{-10pt}
    \scalebox{0.9}{
    \begingroup
    \renewcommand{\arraystretch}{1.0}
    \setlength{\tabcolsep}{2.5pt}
    \begin{tabular}{
        l  
        c  
        c  
        r  
        r  
        r  
    }
    \toprule
    \textbf{Dataset} & \textbf{Source} & \textbf{Dyn.} &
    \textbf{\#Frames} & \textbf{\#Scenes} &
    \textbf{Ratio}  \\
    \cmidrule(lr){1-1}\cmidrule(lr){2-6}
        RealEstate10K~\cite{zhou2018stereo} & Real & \xmark & 10M  & 80K & 1  \\
        DL3DV~\cite{ling2024dl3dv}          & Real & \xmark & 51M  & 10K & 1  \\
        PointOdyssey~\cite{zheng2023pointodyssey} & Syn. & \cmark & 6K   & 131 & 200  \\
        Spring~\cite{mehl2023spring}        & Syn.  & \cmark & 200K & 37  & 500  \\
        Multi-Cam Video~\cite{bai2025recammaster} & Syn. & \cmark & 11M & 13.6K & 1  \\
        DynamicReplica~\cite{karaev2023dynamicstereo} & Real & \cmark & 145K & 484 & 100  \\
        Stereo4D~\cite{jin2025stereo4d}     & Real & \cmark & 15M & 80K & 1  \\
    \bottomrule
    \end{tabular}
    \endgroup}
    \vspace{-5pt}
    \caption{
        Summary of datasets.
        \textbf{Source} indicates whether the dataset is captured from the real world or synthesized.
        \textbf{Dyn}amic specifies whether the scenes are dynamic.
        \textbf{\#Frames} and \textbf{\#Scenes} denote the total number of image frames and unique scenes, respectively.
        \textbf{Ratio} represents the per-scene sampling multiplier used during training for data balancing.
        \vspace{-5pt}
    }
    \label{tab:dataset}
\end{table}

%% file: sec/5_experiment.tex
\section{Ablation: When and Why Elasticity Helps}
\label{sec:abaltion}

\input{floatings/fig_infscale}

Before scaling up the full pretraining pipeline, we perform controlled ablation studies with \model-LVSM at a moderate scale.
These experiments investigate the key algorithmic components added on top of the vanilla LaCT block, including the effects of chunking, anchor update policies, and Fisher estimation.
For this purpose, we start by training the model exclusively on internet stereo videos from Stereo4D~\cite{jin2025stereo4d}, trimmed to a maximum temporal window of 136 frames.
All ablation models use a 12-layer LaCET backbone, trained with a per-GPU batch size of 16 on 8 H100 GPUs, using 32 input and 32 target views, a maximum temporal span of 128 frames, and an image resolution of $128 \times 128$ for 32K steps ($\approx$32B tokens).
We deliberately use these smaller networks so that its long-context performance saturates with a reasonably small number of tokens.
We evaluate on the Stereo4D test set using PSNR~\cite{chan1983hardware}, SSIM~\cite{wang2004image}, and LPIPS~\cite{zhang2018unreasonable}, using 32 randomly sampled views along the trajectory as inputs and averaged over 8 randomly sampled target views per scene.
The results over different settings are aggregated in Table~\ref{tab:ablation}.
More details are available in Appendix~\ref{app:ablation}.

\subsection{Anchor Update Policies}

We analyze how elastic consolidation behaves under different chunking and anchoring configurations.

\boldstartspace{Full-sequence setup (single chunk).}
When the chunk size equals the full sequence length, the model performs exactly one forward pass and one fast-weight update per scene.  
All anchor update policies become equivalent. 
The consolidation term scales with both the update magnitude and the anchor-relative drift, and in the single-chunk regime reduces to a second-order correction $\mathcal{O}(\lambda (\Delta \theta)^2)$ in the update size, which is negligible for small $\lambda$.
    
\boldstartspace{Global anchoring.}
If the anchor weights remain fixed globally, consolidation degenerates into an importance-weighted $\ell_2$ regularizer.
This stabilizes inference-time adaptation, but does not encode temporal continuity beyond the fixed prior, similar to weight decay.

\boldstartspace{Streaming anchoring.}
Under streaming (w/o EMA) update, the anchor is reset to the current fast weights at the beginning of each chunk.  
The consolidation term then only regularizes within-chunk drift, applying adaptive shrinkage to the accumulated fast-weight change. 
This configuration lacks memory consolidation across chunks, making it more prone to overfitting.

\boldstartspace{Streaming-EMA anchoring.}
The non-trivial, genuinely elastic behavior emerges when streaming anchors are combined with EMA updates.  
The consolidation term acts as a low-pass, importance-weighted constraint on the fast-weight trajectory, penalizing cumulative drift relative to an dynamically evolving consolidated anchor weight rather than the instantaneous update.

\input{floatings/fig_steoro4d_main}
\input{floatings/tab_4dnvs}

\subsection{Elasticity Improves Generalization}

As shown in Table~\ref{tab:ablation}, we observe a clear gap between training and test PSNR, i.e., training vs.\ test $\ell_2$, which points to substantial overfitting. 
This generalization gap is reduced by consolidation, suggesting that consolidation improves information transfer across chunks while also suppressing fast-weight drift caused by repeated fully plastic inference-time updates. 
We hypothesize that LaCT-LVSM tends to exploit local pattern shortcuts, effectively memorizing localized cues within its limited fast-weight memory instead of maintaining a more distributed spatiotemporal representation, consistent with similar findings in other efficient architectures~\cite{you2025revealing}. 
We next provide a deeper analysis of what LaCT-LVSM overfits to in practice.

\boldstartspace{Setups.}
Figure~\ref{fig:infscale} examines how LaCT and LaCET behave under different test-time input densities. 
Both models are trained with 32 input images, and we vary the number of input frames at inference on 136-frame Stereo4D clips.
In the discrete-view setting, input and target frames are uniformly sampled across the full span.
In the continuous-view setting, we crop a contiguous sub-sequence (e.g., 40 frames for the 32-in/8-out case) and mask the target frames within that window, reducing the problem to frame interpolation.
Two settings converge when the full 136-frame span is used.

\boldstartspace{LaCET consistently dominates LaCT under sparse inputs.}
When input views are sparse in time and space, the advantages of LaCET are large and systematic across all PSNR/SSIM/LPIPS metrics. 
Both LaCET (4 chunks) and LaCT (4 chunks) degrade sharply as sparsity increases, while LaCT (1 chunk) collapses gracefully, as more activation memory is used to process the full sequence (which is not sustainable for longer sequences).
Nevertheless, smaller chunks remain appealing due to their reduced activation memory footprint, since backpropagation spans fewer samples, making them more suitable for scaling and for real streaming applications.

\boldstartspace{LaCET mitigates camera-pose interpolation shortcuts.}
In the continuous-view regime, LaCET (4 chunks) begins to outperform both LaCT (1 chunk) but still outperforms LaCT (4 chunks).
This behavior reveals that LaCT learns to exploit short-range temporal redundancy rather than learning a true view-conditioned spatial representation.
When input frames are continuous, the task effectively degenerates into a frame interpolation problem. 
The model can simply latch onto neighboring frames in the context window and does not need to perform genuine NVS for 4D representation, i.e., no camera-pose extrapolation or long-range temporal modeling.
\cite{mitchel2026true} made similar observations.
LaCET still improves with more continuous inputs, but the gap between discrete-view and continuous-view performance is substantially smaller.
This indicates that LaCET is less prone to collapsing into an interpolation-only solution and instead preserves the ability to model long-range 4D dynamics.

\input{floatings/fig_dl3dv_compare}
\input{floatings/tab_3dnvs}

\section{Scaling LaCET for Fast Spatial Memory}

\subsection{Pretraining Curriculum}

Based on the controlled studies described above, we default LaCET blocks to (i) the streaming-EMA anchor update policy and (ii) the SI-style importance estimate for empirically better training stability.
We train both the \model-LVSM and \model-LRM variants.
Given compute limitations, we bootstrap the LVSM variant from a DL3DV-pretrained LaCT backbone with a resolution of 128, introduce additional temporal encodings, and continue pretraining it for pose-conditioned 4D reconstruction.
For data scheduling, we employ a long-context curriculum that gradually increases the input resolution (128 $\rightarrow$ 256), the temporal span (128 $\rightarrow$ 256), and dynamic number of input views as training progresses.
Complete implementation details are available in Appendix~\ref{app:scale}.

\subsection{Novel View Synthesis Performance}

For fair comparisons, we report the highest score among (i) our reproduced results, (ii) reported by the authors, and (iii) those reported by the community.
Note that metrics like PSNR are resolution-dependent (e.g., higher resolutions typically produce higher PSNR). 
We adopt the lowest resolution (256$\times$256) for meaningful comparison with baselines.

\boldstartspace{4D Novel View Synthesis.}
Unlike 3D NVS, there is currently no well-established benchmark for feedforward 4D evaluation. 
Existing datasets were originally designed for optimization-based pipelines, and the community has not yet converged on a standard evaluation protocol. 
We use the NVIDIA~\cite{yoon2020novel} benchmarks (with the same evaluation setup in \cite{lin2025movies}) and Steoro4D~\cite{jin2025stereo4d} benchmarks for fair comparison within this regime. 
In Table~\ref{tab:4d_results}, we show our method outperforms existing approaches evaluated at similar resolutions. 
In particular, on Stereo4D our model achieves clear improvements over prior rendering-based methods across all metrics.
On the NVIDIA benchmark, our method achieves the best performance among feed-forward approaches at $256\times256$ resolution, and approaches the performance of the strongest optimization-based methods, which require per-scene test-time optimization.
These results suggest that the proposed LaCET effectively benefits dynamic scene modeling, where maintaining consistent spatial information across time becomes critical.

\boldstartspace{3D Novel View Synthesis.}
We use the DL3DV-140 benchmark~\cite{ling2024dl3dv} for evaluation.
Since evaluation metrics scale with resolution, we adopt the minimal 256 resolution to ensure fair comparison across both categories.
In Table~\ref{tab:3d_results}, we show our method delivers performance comparable to existing approaches evaluated at similar resolutions, demonstrating that the proposed LaCET blocks preserve strong capability on static scenes where spatial memory is less critical.

%% file: floatings/fig_infscale.tex
\begin{figure*}[!t]
\centering
    \begin{subfigure}[t]{.95\textwidth}
        \centering
        \includegraphics[width=1.02\textwidth]{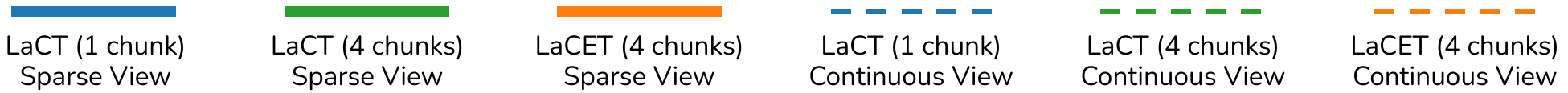}
        \vspace*{-15pt}
    \end{subfigure}
    ~
    \begin{subfigure}[t]{.32\textwidth}
        \centering
        \includegraphics[width=1.02\textwidth]{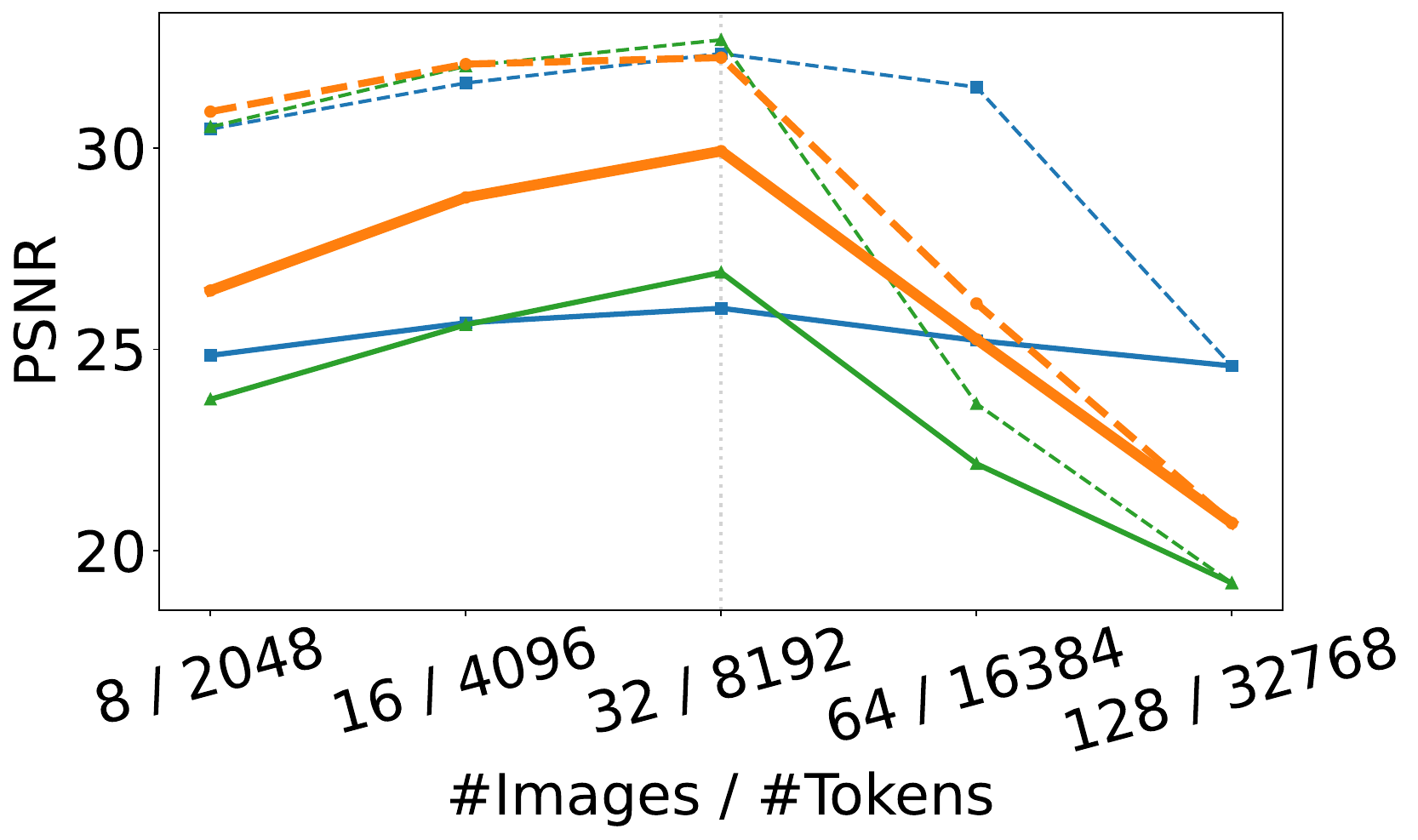}
        \vspace*{-15pt}
        \caption{PSNR vs. \#input imgs / \#tokens.}
    \end{subfigure}
    ~
    \hspace{-5pt}
    \begin{subfigure}[t]{.32\textwidth}
        \centering
        \includegraphics[width=1.02\textwidth]{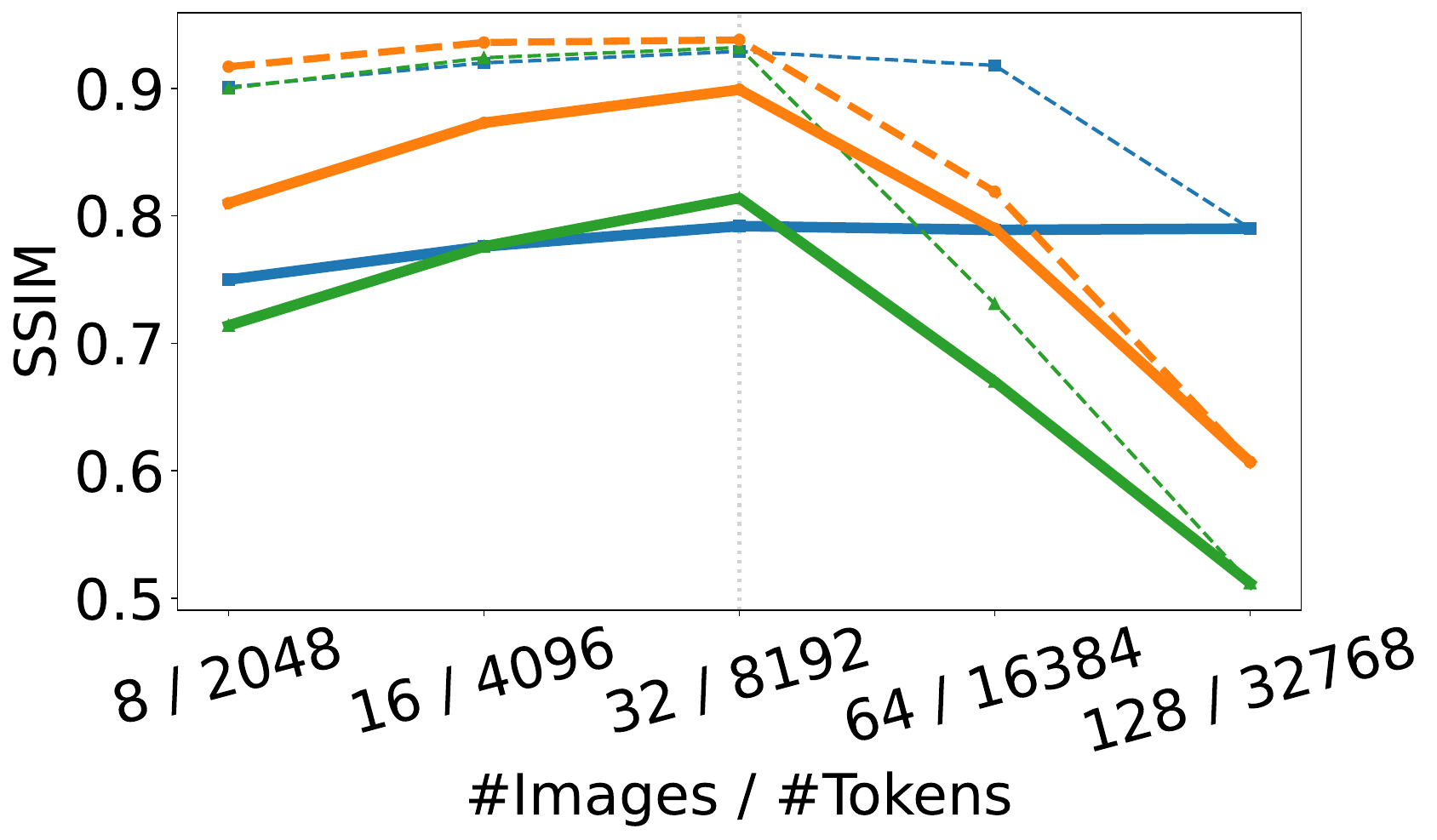}
        \vspace*{-15pt}
        \caption{SSIM vs. \#input imgs / \#tokens.}
    \end{subfigure}
    ~
    \hspace{-5pt}
    \begin{subfigure}[t]{.32\textwidth}
        \centering
        \includegraphics[width=1.02\textwidth]{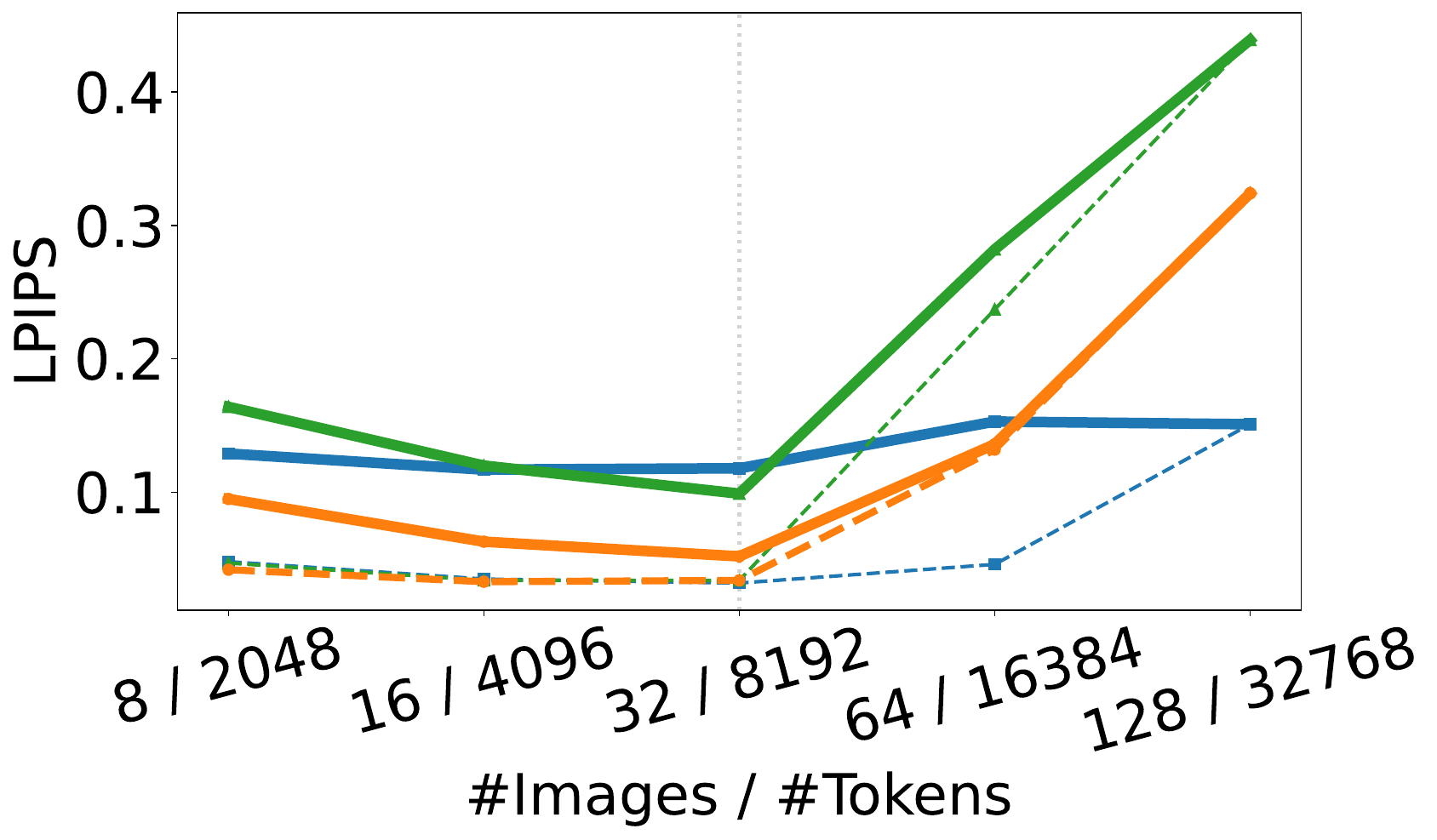}
        \vspace*{-15pt}
        \caption{LPIPS vs. \#input imgs / \#tokens.}
    \end{subfigure}
    \vspace{-5pt}
    \caption{
    \textbf{Test-time scaling curves.} 
    Shown are PSNR/SSIM/LPIPS of LaCT (1/4 chunks) and LaCET (4 chunks; streaming-ema), trained with 32 images (vertical line) and evaluated with varying numbers of input images. Each point uses a 136-frame Stereo4D clip.
    \textbf{For sparse views}, input and target frames are randomly sampled across the long full span.
    For \textbf{continuous views}, we select a contiguous sub-sequence (e.g., 40 frames for 32-in/8-out) and randomly mask the target frames inside it for the model to predict, reducing to frame interpolation.
    \vspace{-10pt}
    }
    \label{fig:infscale}
\end{figure*}

%% file: floatings/fig_steoro4d_main.tex
\begin{figure*}[t!]
    \centering
    \includegraphics[width=1.0\linewidth]{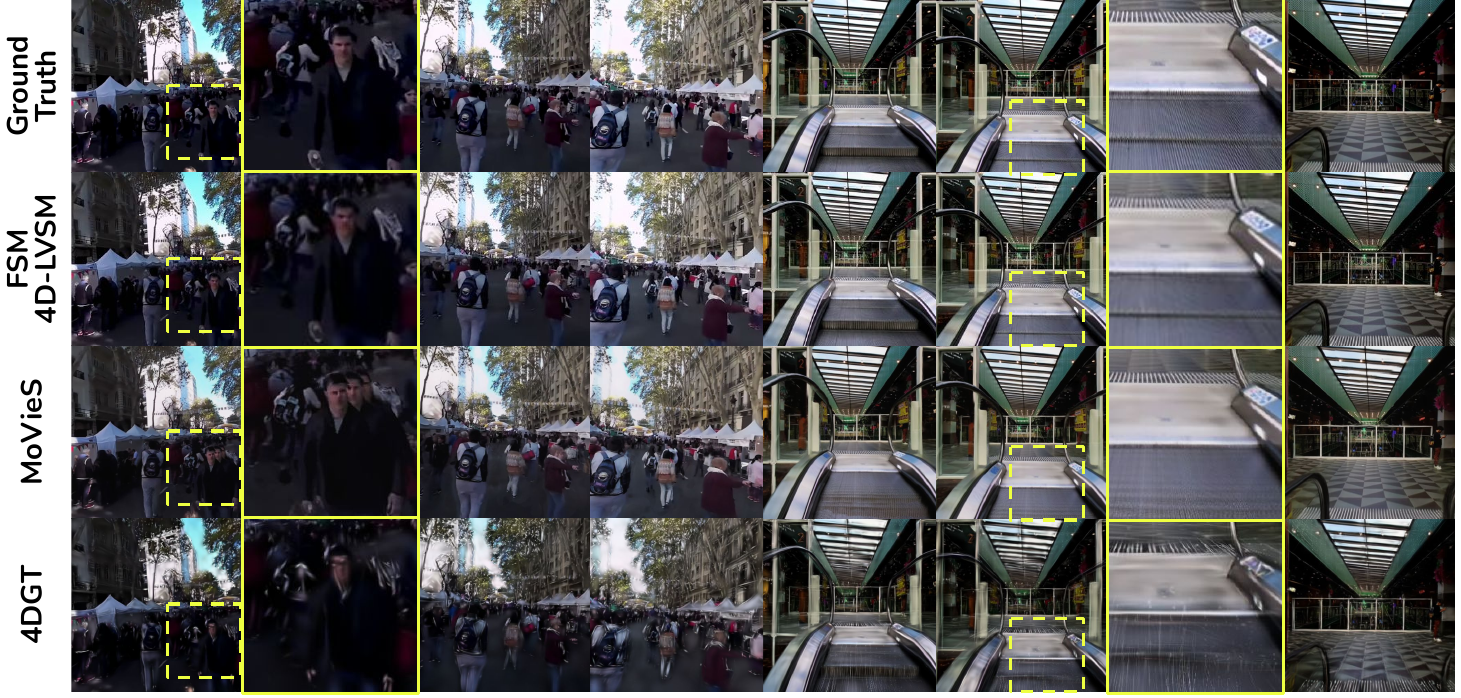}
    \vspace{-15pt}
    \caption{Qualitative comparison on Steoro4D test set. Note that for MoVieS we use  a higher default resolution (504 $\times$ 504). \vspace{-5pt}}
    \label{fig:steoro4d}
\end{figure*}

%% file: floatings/tab_4dnvs.tex
\begin{table*}[!t]
    \centering
    \hspace{-10pt}
    \scalebox{0.95}{
    \begingroup
    \renewcommand{\arraystretch}{0.9}
    \setlength{\tabcolsep}{11pt}
    \begin{threeparttable}
    \begin{tabular}{lcccccccc}
    \toprule
    \multirow{2}{*}{\textbf{Model}} & \multicolumn{4}{c}{\textbf{Stereo4D}~\cite{jin2025stereo4d}}                           & \multicolumn{4}{c}{\textbf{NVIDIA}~\cite{yoon2020novel}}                            \\
    \cmidrule(lr){2-5}\cmidrule(lr){6-9}
                                    & \textbf{Resolution} & \textbf{PSNR}$^\uparrow$ & \textbf{LPIPS}$^\downarrow$ & \textbf{SSIM}$^\uparrow$ & \textbf{Resolution} & \textbf{PSNR}$^\uparrow$ & \textbf{LPIPS}$^\downarrow$ & \textbf{SSIM}$^\uparrow$ \\
    \cmidrule(lr){1-1}\cmidrule(lr){2-5}\cmidrule(lr){6-9}
    \rowcolor[HTML]{ffffcc}
    \multicolumn{9}{c}{\textit{Optimization-based}} \\
    SoM~\cite{wang2025shape}           & \multicolumn{4}{c}{\textcolor{gray}{------ OOT$^{\star}$ ------}} & 379 $\times$ 672 & 15.30 & 0.509 & 0.317 \\
    MoSca~\cite{lei2025mosca}         & \multicolumn{4}{c}{\textcolor{gray}{------ OOT$^{\star}$ ------}} & 379 $\times$ 672 & 21.45 & 0.265 & 0.712 \\
    \cmidrule(lr){1-1}\cmidrule(lr){2-5}\cmidrule(lr){6-9}
    \rowcolor[HTML]{ffffcc}
    \multicolumn{9}{c}{\textit{Rendering-based}} \\
    L4GM~\cite{ren2024l4gm}          & \multicolumn{4}{c}{\textcolor{gray}{------ OOT$^{\dagger}$ ------}}  & 256 $\times$ 256 & 10.07 & 0.587 & 0.235 \\
    4DGT~\cite{xu20254dgt}          & 504 $\times$ 504 &  24.62 &  0.102 & 0.785     & 504 $\times$ 504 &  14.13 &  0.640 & 0.131    \\
    MoVieS~\cite{lin2025movies}        & 504 $\times$ 504 &  27.19 &  0.114 & 0.888 & 379 $\times$ 672 & 19.16 & 0.315 & 0.514 \\
    \textbf{\model-LRM} & 256 $\times$ 256 &  27.29 &  0.147 & 0.876 & 256 $\times$ 256 & 20.17 & 0.337 & 0.567  \\
    \textbf{\model-LVSM} & 256 $\times$ 256 &  \textbf{32.16	} &  \textbf{0.043} & \textbf{	0.931} & 256 $\times$ 256 & \textbf{23.90} & \textbf{0.105} & \textbf{0.747}  \\
    \bottomrule
    \end{tabular}
    \begin{tablenotes}
    \item[$\star$] \textit{SoM takes around 10min per scene and MoSca takes around 45min per scene.}
    \item[$\dagger$] \textit{L4GM requires multi-view diffusion as prior.}
    \end{tablenotes}
    \end{threeparttable}
    \endgroup}
    \vspace{-5pt}
    \caption{
        \textbf{4D NVS Results.}
         Metrics are resolution-dependent (e.g., higher resolutions typically produce higher PSNR). 
         We adopt the lowest resolution for meaningful comparison with baselines.
         Steoro4D test set contains 7109 scenes, which is out of time (OOT) for some methods.
         \vspace{-10pt}
    }
    \label{tab:4d_results}
\end{table*}

%% file: floatings/fig_dl3dv_compare.tex
\begin{figure*}[t!]
    \centering
    \includegraphics[width=1.0\linewidth]{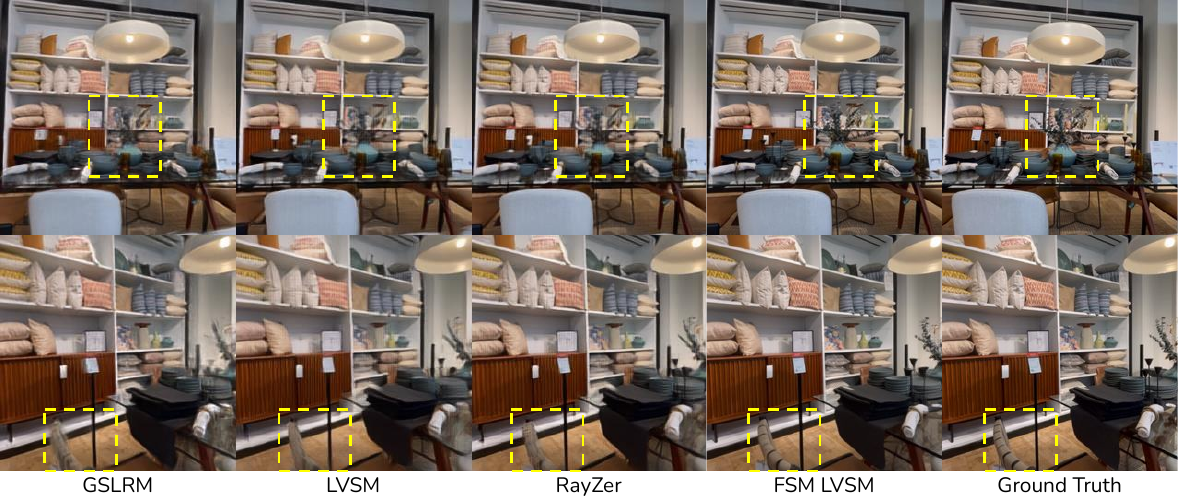}
    \vspace{-20pt}
    \caption{Qualitative comparison on DL3DV benchmark. \vspace{-10pt}}
    \label{fig:dl3dv}
\end{figure*}

%% file: floatings/tab_3dnvs.tex
\begin{table}
    \centering
    \hspace{-10pt}
    \scalebox{0.9}{
    \begingroup
    \renewcommand{\arraystretch}{0.9}
    \setlength{\tabcolsep}{6pt}
    \begin{threeparttable}
    \begin{tabular}{lcccc}
    \toprule
    \multirow{2}{*}{\textbf{Model}} & \multicolumn{4}{c}{\textbf{DL3DV}~\cite{ling2024dl3dv}}                             \\
    \cmidrule(lr){2-5}
                                    & \textbf{Resolution} & \textbf{PSNR}$^\uparrow$ & \textbf{LPIPS}$^\downarrow$ & \textbf{SSIM}$^\uparrow$ \\
    \cmidrule(lr){1-1}\cmidrule(lr){2-5}
    \rowcolor[HTML]{ffffcc}
    \multicolumn{5}{c}{\textit{Static Models}} \\
    DepthSplat~\cite{xu2025depthsplat}                      & 512 × 448     & 17.81         & 0.356          & 0.596         \\
    GS-LRM~\cite{zhang2024gslrm}                          & 256 × 256     & 23.02         & 0.266          & 0.705         \\
    LVSM~\cite{jin2025lvsm}                            & 256 × 256     & 23.10         & 0.257          & 0.703         \\
    RayZer$^\dagger$~\cite{jiang2025rayzer}                         & 256 × 256     & 23.72         & 0.222          & 0.733         \\
    LongLRM~\cite{ziwen2025long}                         & 540 × 960     & 24.10         & 0.254          & 0.783         \\
    tttLRM~\cite{wang2026tttlrm}                         & 540 × 960     & 25.07         & 0.215          & 0.822         \\
    tttLVSM~\cite{zhang2025test}                       & 540 × 960     & \textbf{26.90}         & 0.185          & 0.837         \\
    \textbf{\model-LRM}                   & 256 × 256     & 23.59	         & 0.206	        & 0.766          \\
    \textbf{\model-LVSM}                   & 256 × 256     & \uline{26.69}         & \textbf{0.091}          & \textbf{0.846}         \\
    \cmidrule(lr){1-1}\cmidrule(lr){2-5}
    \rowcolor[HTML]{ffffcc}
    \multicolumn{5}{c}{\textit{Dynamic Models}} \\
    \textbf{\model-LRM}                   & 256 × 256     & 21.89	         & 0.314          & 	0.692         \\
    \textbf{\model-LVSM}                   & 256 × 256     & 24.61	       & 0.118	         & 0.787         \\
    \bottomrule
    \end{tabular}
    \begin{tablenotes}
    \item[$\dagger$] \textit{RayZer ignores input poses and uses target reference images instead, placing it somewhere between pose-conditioned and fully pose-free approaches.}
    \end{tablenotes}
    \end{threeparttable}
    \endgroup}
    \vspace{-5pt}
    \caption{
        \textbf{3D NVS Results.}
         Metrics are resolution-dependent (e.g., higher resolutions typically produce higher PSNR). We adopt the lowest resolution for meaningful comparison with baselines.
        \vspace{-10pt}
    }
    \label{tab:3d_results}
\end{table}

%% file: sec/2_related.tex
\section{Related Work}
\label{sec:background}

\boldstart{Fast Weights and Test-Time Training (TTT).}
Recently, many sequence models have been reformulated under the lens of inference-time learning or regression, which interprets the recurrent update of model states as a form of online learning~\cite{liu2025longhorn} from context~\cite{von2023transformers,dherin2025learning,behrouz2025atlas}.
This view commonly connects modern sequence models to the long-standing notion of fast weights~\cite{schmidhuber1992learning}, i.e., parameters that evolve in-context at each timestep to capture short-term associations.
Fast-weight mechanisms thus act as associative memories~\cite{bietti2023birth,ramsauer2021hopfield}, balancing retention and adaptation through architectures such as DeltaNet~\cite{schlag2021linear,yang2024parallelizing}.
Recently, Test-Time Training (TTT) extends fast-weight adaptation to general neural components that update online using self-supervised signals~\cite{sun2025learning,wang2025test}.
Recent works explore specialized test-time optimizers~\cite{behrouz2025titans,karami2025lattice} and online learning objectives~\cite{behrouz2025s}, with applications in video generation, 3D reconstruction, and beyond~\cite{dalal2025one,chen2026ttt3r,zhang2025test}.
However, na\"ive TTT remains bottlenecked by poor hardware utilization, limited state capacity, and unstable long-horizon dynamics~\cite{tandon2025end}. 
Large-Chunk Test-Time Training (LaCT) improves this paradigm by enabling efficient in-forward fast-weight updates over larger contexts~\cite{zhang2025test,liu2026test}. 
Still, LaCT relies on fully plastic fast-weight dynamics, which can lead to overfitting and catastrophic forgetting over long sequences. 
This work addresses this issue with Elastic TTT, which stabilizes fast-weight adaptation by introducing additional elasticity across chunks.

\boldstart{Large Rendering-Based Reconstruction Models.} 
Large Reconstruction Models (LRMs) have recently emerged as a unified framework for producing view-consistent 3D reconstructions.
Trained on massive 3D and 4D datasets, these models leverage triplane-based NeRFs~\cite{li2024instant3d,hong2024lrm,wang2024pflrm,jiang2025real3d} or Gaussian Splatting~\cite{zhang2024gslrm,xie2024lrm,ziwen2025long,ziwen2025longplus,wang2026tttlrm} to encode strong priors over shape and appearance, achieving high-quality reconstruction from only a few posed views.
In the 4D setting, similarly, existing LRMs still rely heavily on geometric supervision to maintain rendering consistency, typically requiring posed inputs together with explicit Gaussian primitives~\cite{ren2024l4gm,ma20254d,xu20254dgt,yang2025storm,liang2025feedforward,lin2025movies}.
More recently, Large View Synthesis Models (LVSMs) have begun to relax these geometric constraints, achieving high-quality view synthesis without explicit geometric representations~\cite{jin2025lvsm,zhang2025test,kim2026scaling} and, in some cases, supporting self-supervised autoencoding reconstruction~\cite{jiang2025rayzer,chen2026wildrayzer,mitchel2026true}.
Our work follows this direction by developing a fast 4D reconstruction model that learns scene-level spatiotemporal representations, and by instantiating it both with and without minimal geometric priors.
A parallel line of research explores feed-forward, geometry-centric reconstruction models~\cite{wang2024dust3r,tang2024mv,wang2025vggt,yang2025fast3r,zhang2026loger} through large-scale training.
These methods have inspired several 4D counterparts that estimate dynamic geometry or camera poses without supporting novel view-time synthesis~\cite{zhang2024monst3r,wang2025continuous,feng2025st4rtrack,zhuo2025streaming,zhou2026page,chen2026ttt3r}.
This work departs from explicit geometric reconstruction and instead treats novel view-time synthesis as the core objective of 4D representation learning, following prior work~\cite{zhang2025test,liu2026test,kim2026scaling} that has used this task as the primary task for training, evaluation, and scaling-law studies of model architecture.

%% file: sec/6_conclusion.tex
\section{Conclusion and Limitations}

\boldstartspace{Scaling to Longer Sequences.}
LaCET enables fast inference-time adaptation for high-quality rendering from, in principle, arbitrarily long sequences in a single forward pass, where activation memory is no longer the bottleneck. 
However, due to limitations in licensable training data and suitable benchmarks, as well as our compute budget, we focus in this work on architectural advances rather than training and scaling a model that fully realizes the method's potential.

\boldstartspace{Pose Estimation in Dynamic Scenes.}
Recently, several works have explored 3D reconstruction from unposed images~\cite{wang2024pflrm,jiang2025rayzer,mitchel2026true}.
However, jointly estimating camera intrinsics and poses in dynamic scenes, where both camera motion and scene dynamics are present remains challenging.
In this work, we assume posed input images and do not treat unposed reconstruction as a primary target.

\boldstartspace{Geometrically Faithful 4D Reconstruction.}
While NVS is a key task for spatial intelligence, solving it does not by itself ensure geometric faithfulness or temporally consistent motion.
Accurate 4D geometry requires additional constraints and evaluation protocols beyond view synthesis quality.
There is ongoing debate in the community over whether explicit geometric supervision is necessary, or whether rendering-based supervision alone is sufficient for learning geometrically faithful representations.
In this work, we deliberately focus on the architectural aspects of this problem.
While LaCET reduces the tendency of the model to interpolate nearby context frames instead of performing true NVS, this behavior does not fully disappear under rendering-only supervision. 
We expect that incorporating additional geometric supervision, e.g., depth, correspondence, multi-view consistency, or motion cues such as optical flow, could further mitigate this issue, and we leave this direction to future work.

\boldstartspace{Acknowledgment.}
The authors would like to thank Zefan Cai, Xuweiyi Chen, Yinpei Dai, Yilun Du, Chenguo Lin, Freda Shi, Hao Tan, Zeyuan Yang, and Tianyuan Zhang for their insightful discussions.

%% file: sec/a1_method.tex
\section{Implementation and Training Details}
\label{app:method}

\subsection{Data Pre-processing}
\label{app:preprocess}

For each training sample, we load a video clip together with per-frame camera metadata, including intrinsics and world-to-camera poses. 
We first sample a temporal window from the full clip, then randomly select input and target frames within that window. 
For each selected frame, we extract the RGB image from the video, convert the stored world-to-camera matrix to camera-to-world form, and collect the corresponding intrinsic parameters. 
The image is resized and cropped to the target resolution, while the intrinsics are updated accordingly. 
All images are converted to RGB and normalized to tensors.
The frame timestamp is taken from the frame index, then normalized within the sampled clip segment with linear rescale. 
This preserves relative temporal ordering while keeping timestamps in a fixed range across videos of different lengths. 
We further normalize camera poses at the scene level by centering them with respect to the mean pose.

\subsection{Algorithm and Model Architecture}

For the elastic test-time training algorithm, we use $\alpha_\text{ewc}=0.5$, $\beta_\text{ewc}=0.5$ and $\lambda_\text{ewc}=0.5$ after grid search.
Each block uses a model dimension of 768 and the fast-weight module is implemented as a single-head SwiGLU MLP with a hidden dimension of 1536. 
The window attention module contains 12 heads with a head dimension of 64 and applies QK-Norm~\cite{henry2020query}. 
The feed-forward network uses an intermediate hidden dimension of 3072. 
Both the tokenization and decoder layer are linear projections, with a sigmoid applied at the decoder. 
During both training and inference, the update operation is applied to all input tokens, and the fast weights are subsequently used to process the target tokens.
All model variants in this paper use the same LaCET block configuration and update rule. 

\subsection{Ablation Study Settings}
\label{app:ablation}

For the ablation study in Sec.~\ref{sec:abaltion}, we adopt a controlled configuration with 12 LaCET blocks.

\boldstartspace{Data usage.}
We conducted all experiments on Stereo4D~\cite{jin2025stereo4d}, a large dataset containing diverse camera trajectories and both static and dynamic object motion, which makes it well suited for modeling 4D scenes.
We followed its official train-test splits.

\boldstartspace{Training details.}
For ablation study, we train with with 32 input views and 32 novel views at $128\times128$ resolution for 32K steps. 
During training, we first sample a window of 128 consecutive frames, then randomly select 64 frames, from which 32 are used as input and the remaining 32 as target views. 
The detailed training configuration is provided in Table~\ref{tab:config}. 
All experiments are trained on 8 H100 GPUs.

\subsection{Full-Scale Pre-training Settings}
\label{app:scale}

\boldstartspace{Data usage.}
To scale up the model capacity, we train the complete \model model on a large collection of both synthetic and real data in Table~\ref{tab:dataset}. 

\boldstartspace{Training details.}
We first pre-train our model at $128\times128$ resolution for 80K steps, and then fine-tune it at $256\times256$ resolution for an additional 10k steps. 
All training configurations use 32 context frames and 32 target frames, sampled from a window of 128 consecutive frames. 
Detailed training settings are provided in Table~\ref{tab:config}. 
Both training stages are done with 64 H100 GPUs. 

\input{floatings/tab_params}

%% file: floatings/tab_params.tex
\begin{table}[!t]
    \centering
    \hspace{-10pt}
    \scalebox{0.85}{
    \begingroup
    \renewcommand{\arraystretch}{0.8}
    \setlength{\tabcolsep}{2.5pt}
    \begin{tabular}{ccccc}
    \toprule
    \textbf{Config}    & \textbf{Ablation} & \textbf{Base} & \textbf{Resolution} & \textbf{Multi-Length} \\
    \textbf{Parameters} & \textbf{Training} & \textbf{Training} & \textbf{Scaling} & \textbf{Fine-tuning} \\
    \midrule
    \#layers           & 12       & 24      & 24            & 24         \\
    \#input frames     & 32       & 32      & 32            & 12-64      \\
    \#target frames    & 32       & 32      & 32            & 32         \\
    resolution         & 128      & 128     & 256           & 256        \\
    temporal window    & 128       & 128     & 256           & 256        \\
    optimizer          & Adam     & Adam    & Adam          & Adam       \\
    beta 1              & 0.9      & 0.9     & 0.9           & 0.9        \\
    beta 2              & 0.95     & 0.95    & 0.95          & 0.95       \\
    weight decay       & 0.05      & 0.05     & 0.05           & 0.05        \\
    learning rate      & 2e-4     & 1e-4    & 5e-5          & 1e-4       \\
    lambda L2          & 1.0      & 1.0     & 1.0           & 1.0        \\
    lambda LPIPS       & 0.5      & 0.5     & 0.5           & 0.5       \\
    batch size per gpu & 16       & 16      & 4             & 4          \\
    \#gpus             & 8        & 64      & 64            & 64        \\
    L2 warmup          & 1000     & 2500    & 500          & 0          \\
    warmup steps       & 1000     & 2500    & 1000          & 0          \\
    total steps        & 32000    & 80000   & 20000         & 20000      \\
    \bottomrule
    \end{tabular}
    \endgroup}
    \caption{
        Summary of configurations across ablation studies, base training, resolution scaling, and variable-length fine-tuneing.
    }
    \label{tab:config}
\end{table}

%% file: sec/a2_results.tex
\section{Addendum to Results and Discussions}
\label{app:experiment-details}

\subsection{Batch Inference}

Unlike standard inference, LaCET modifies the model state during inference through fast-weight updates.
When the inference batch size is greater than 1, updates from all examples in the batch are averaged (or accumulated) and applied once per chunk.
Consequently, batch size directly affects the adaptation dynamics rather than merely the throughput, which is a distinctive property of test-time-training architectures that makes batched inference behave similarly to dynamic evaluation~\cite{krause2018dynamic} or few-shot adaptation.
Empirically, we found the effect to be minimal (Table~\ref{tab:ablation}); nevertheless, we fix the inference batch size to 1 in all subsequent experiments.

\input{floatings/tab_ablation_app}

\subsection{LVSM-style Decoder vs. LRM-style Decoder}

We provide additional side-by-side ablations comparing LVSM-style vs. LRM-style decoders.

\boldstartspace{LVSM-style decoder.}
In a typical LVSM-style, no explicit scene representation is used in modeling.
We use a shallow image-token decoder to reconstruct pixel patches from token embeddings. 
Specifically, for each token, we first apply layer normalization, followed by a linear projection from the token dimension to $3p^2$, where $p$ denotes the patch size. 
The resulting vector is interpreted as the flattened RGB values of the reconstructed patch. 
A sigmoid activation is applied at the output to bound predictions to ($[0,1]$), matching normalized pixel space.

\boldstartspace{LRM-style decoder.}
With explicit 4D representation, e.g., 4DGS~\cite{yang2024realtime}, we implement a model following 4D-LRM~\cite{ma20254d} and tttLRM~\cite{wang2026tttlrm}.
To adapt large-chunk TTT for explicit GS modeling, we query the fast weights for a set of virtual view planes for 4DGS and used the input views as the virtual views.
We adopt pixel-aligned Gaussian rendering, giving $V \times H \times W$ Gaussians, each with $\dim_\mathrm{4DGS} = 20$.
From each decoded 4D Gaussian parameter $\mathbf{g}\in \mathbb{R}^{20}$, we split the 4-channel space-time vector $(\mathbf{g}_\mathrm{x}, \mathbf{g}_\mathrm{y}, \mathbf{g}_\mathrm{z}, \mathbf{g}_\mathrm{t})$, retain the time $\mu_t = \mathbf{g}_\mathrm{t}$, and normalize the $\mathrm{xyz}$ features to a scalar distance $\delta$.
We strictly followed the tile-based rasterization pipeline introduced in 4D-LRM with deferred backpropagation during rendering to reduce GPU memory consumption.
Following the setup in \cite{zhang2024gslrm}, we set $\delta_\mathrm{near} = 0$ and $\delta_\mathrm{far} = 400$. 

\boldstartspace{Results.}
We find that monocular video training leads to substantially less overfitting to camera interpolation, although convergence becomes markedly slower. 
With the same number of training steps as in Table~\ref{tab:ablation_app}, LVSM-style decoding performs better than explicit 4DGS modeling. 
We hypothesize that, while explicit scene representations may offer stronger generalization and robustness, they are also considerably harder to optimize and more computationally expensive.

\subsection{Explicit Temporal Encoding vs. RoPE}

\boldstartspace{Timestamp maps as time conditioning.}
Following 4D-LRM~\cite{ma20254d}, we represent temporal conditioning with a timestamp map that stores the normalized time of each frame. 
For each view, we concatenate this timestamp map with the RGB image and the Plücker ray map along the channel dimension to form a 10-channel feature map. This per-pixel representation encodes both spatial and temporal cues, enabling the model to distinguish not only between camera views but also between different points in time.

\boldstartspace{RoPE-style time conditioning.}
As an alternative to explicit temporal conditioning, we encode frame time directly in the latent tokens using rotary positional embeddings (RoPE).
Each frame is assigned a normalized timestamp, which determines a sinusoidal rotation applied to the first few channels of every token from that frame. 
Since all tokens within a view share the same temporal rotation, the encoding captures frame identity at the view level without entangling time with local spatial layout. This provides a parameter-free and computationally efficient alternative to explicit temporal conditioning.

\boldstartspace{Results.}
We find that using RoPE leads to slower convergence. 
With the same number of training steps as in Table~\ref{tab:ablation_app}, explicit temporal encoding performs better than RoPE. 
We hypothesize that explicit time conditioning provides a stronger and more direct optimization signal, whereas RoPE injects temporal information more implicitly through feature-space rotations, making it harder for the model to learn to use temporal cues efficiently under a limited training budget.

\subsection{Addition Qualitative Results}

We provide additional results in Figures~\ref{fig:nvidia}, ~\ref{fig:stereo4d_app} and~\ref{fig:dl3dv_app}.

\subsection{Failure Cases and Analysis}

Figure~\ref{fig:failure} illustrates a typical failure case. 
Under large camera or view interpolation, the model may fail to update subject motion consistently, instead preserving stale gestures or partial motion patterns from neighboring frames. 
The results also exhibit ghosting artifacts, with residual duplicated structures around moving limbs and bodies.
This suggests that the model still struggles to maintain accurate space-time correspondence and motion consistency when extrapolating across more challenging viewpoints.

\input{floatings/fig_steoro4d_app_compare}
\input{floatings/fig_steoro4d_app}
\input{floatings/fig_failure}
\input{floatings/fig_nvidia}
\input{floatings/fig_dl3dv_app}

%% file: floatings/tab_ablation_app.tex
\begin{table*}[!t]
    \centering
    \hspace{-12pt}
    \scalebox{0.95}{
    \begingroup
    \renewcommand{\arraystretch}{1.0}
    \setlength{\tabcolsep}{8pt}
    \begin{tabular}{lcccccccc}
    \toprule
    \multirow{2}{*}{\textbf{Model}} & \multicolumn{4}{c}{\textbf{DL3DV}~\cite{ling2024dl3dv}}                           & \multicolumn{4}{c}{\textbf{Stereo4D}~\cite{jin2025stereo4d}}                            \\
    \cmidrule(lr){2-5}\cmidrule(lr){6-9}
                                    & \textbf{Res.} & \textbf{PSNR}$^\uparrow$ & \textbf{LPIPS}$^\downarrow$ & \textbf{SSIM}$^\uparrow$ & \textbf{Res.} & \textbf{PSNR}$^\uparrow$ & \textbf{LPIPS}$^\downarrow$ & \textbf{SSIM}$^\uparrow$ \\
    \cmidrule(lr){1-1}\cmidrule(lr){2-5}\cmidrule(lr){6-9}
    FSM-LRM & 128 $\times$ 128 & 20.99 & 0.243 & 0.683 & 128 $\times$ 128 & 28.19 & 0.097 & 0.897 \\
    FSM-LVSM    & 128 $\times$ 128 & \textbf{21.25} & \textbf{0.169} & \textbf{0.655} & 128 $\times$ 128 & \textbf{31.06} & \textbf{0.041} & \textbf{0.931} \\
    FSM-LVSM (w/ RoPE) & 128 $\times$ 128 & 20.75 & 0.237 & 0.680 & 128 $\times$ 128 & 30.54 & 0.059 & 0.922  \\
    \bottomrule
    \end{tabular}
    \endgroup}
    \caption{
        Additional ablation study results on (i) side-by-side comparison of LVSM-style decoder vs. LRM-style decoder and (ii) using explicit temporal channel vs. using RoPE.
    }
    \label{tab:ablation_app}
\end{table*}

%% file: floatings/fig_steoro4d_app_compare.tex
\begin{figure*}[t!]
    \centering
    \includegraphics[width=1.0\linewidth]{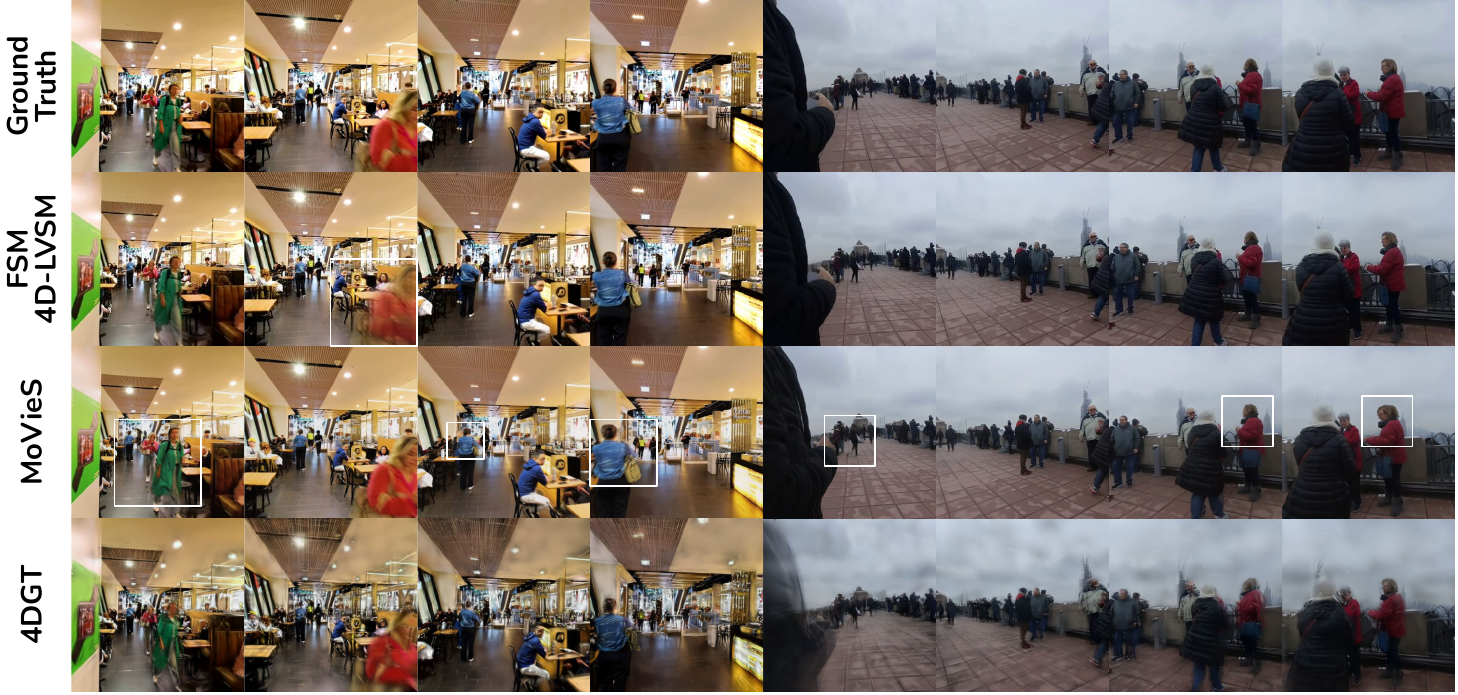}
    \vspace{-15pt}
    \caption{Additional comparison on Steoro4D test set. Note that for MoVieS we use  a higher default resolution (504 $\times$ 504). \vspace{-5pt}}
    \label{fig:steoro4d_app_2}
\end{figure*}

%% file: floatings/fig_steoro4d_app.tex
\begin{figure*}[htbp]
    \centering
    \includegraphics[width=1.0\linewidth]{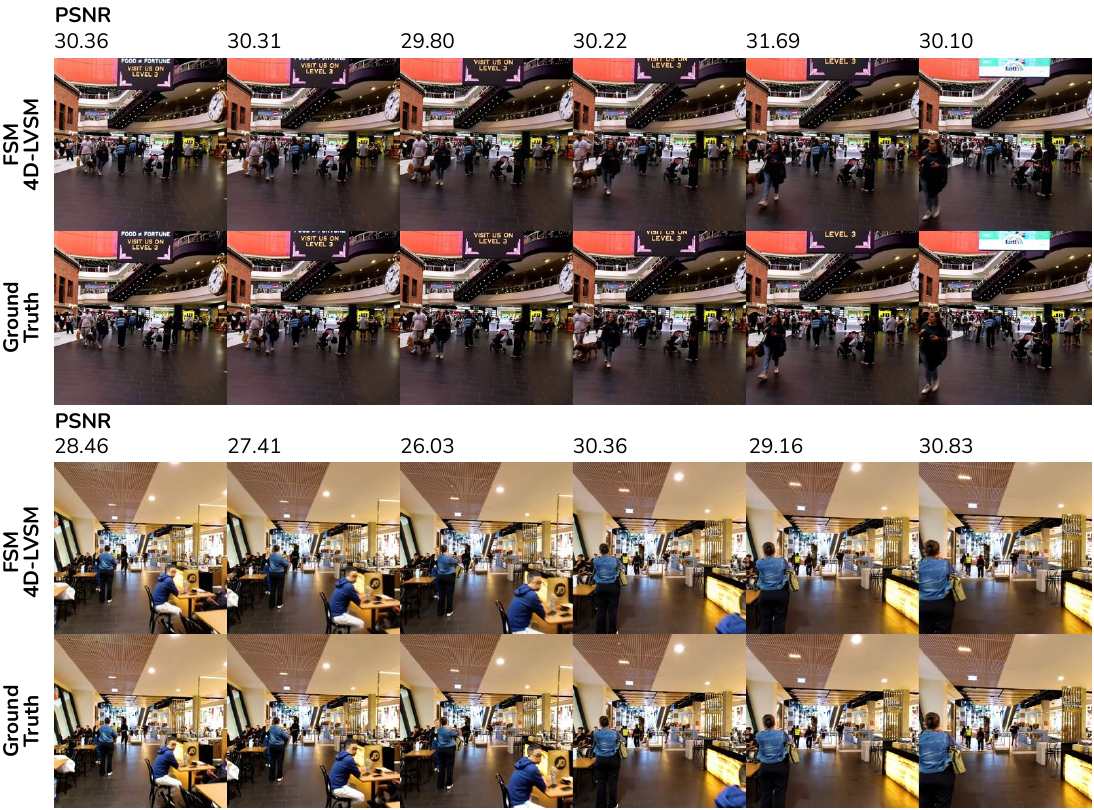}
    \vspace{-15pt}
    \caption{Qualitative examples on Steoro4D test set. \vspace{-5pt}}
    \label{fig:stereo4d_app}
\end{figure*}

%% file: floatings/fig_failure.tex
\begin{figure*}[htbp]
    \centering
    \includegraphics[width=1.0\linewidth]{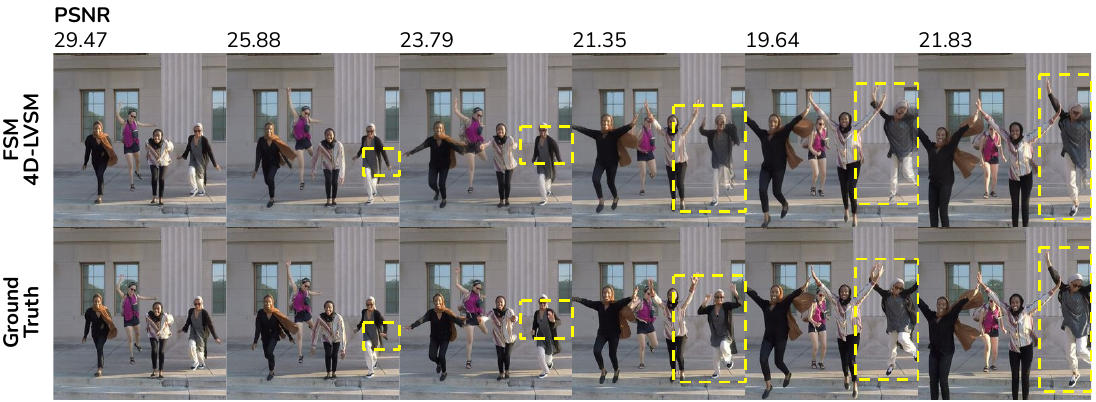}
    \caption{Qualitative failure example. \vspace{-5pt}}
    \label{fig:failure}
\end{figure*}

%% file: floatings/fig_nvidia.tex
\begin{figure*}[htbp]
    \centering
    \includegraphics[width=1.0\linewidth]{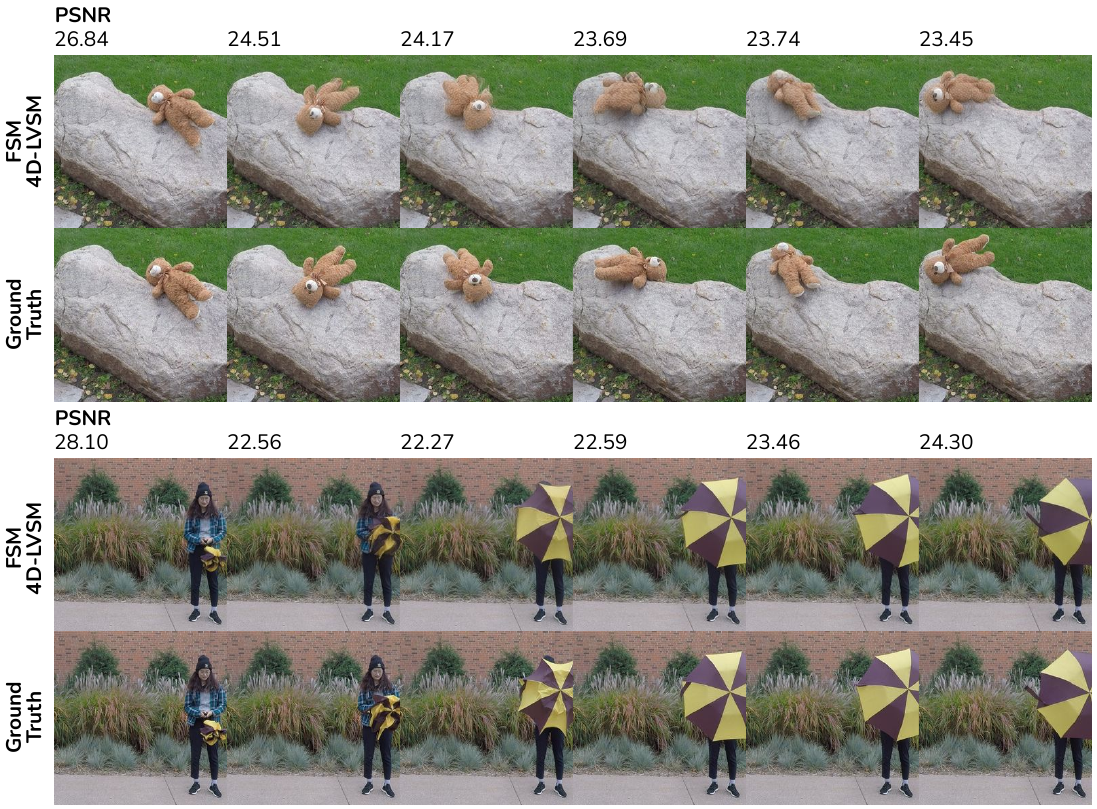}
    \vspace{-15pt}
    \caption{Qualitative results on NVIDIA benchmark. \vspace{-5pt}}
    \label{fig:nvidia}
\end{figure*}

%% file: floatings/fig_dl3dv_app.tex
\begin{figure*}[htbp]
    \centering
    \includegraphics[width=1.0\linewidth]{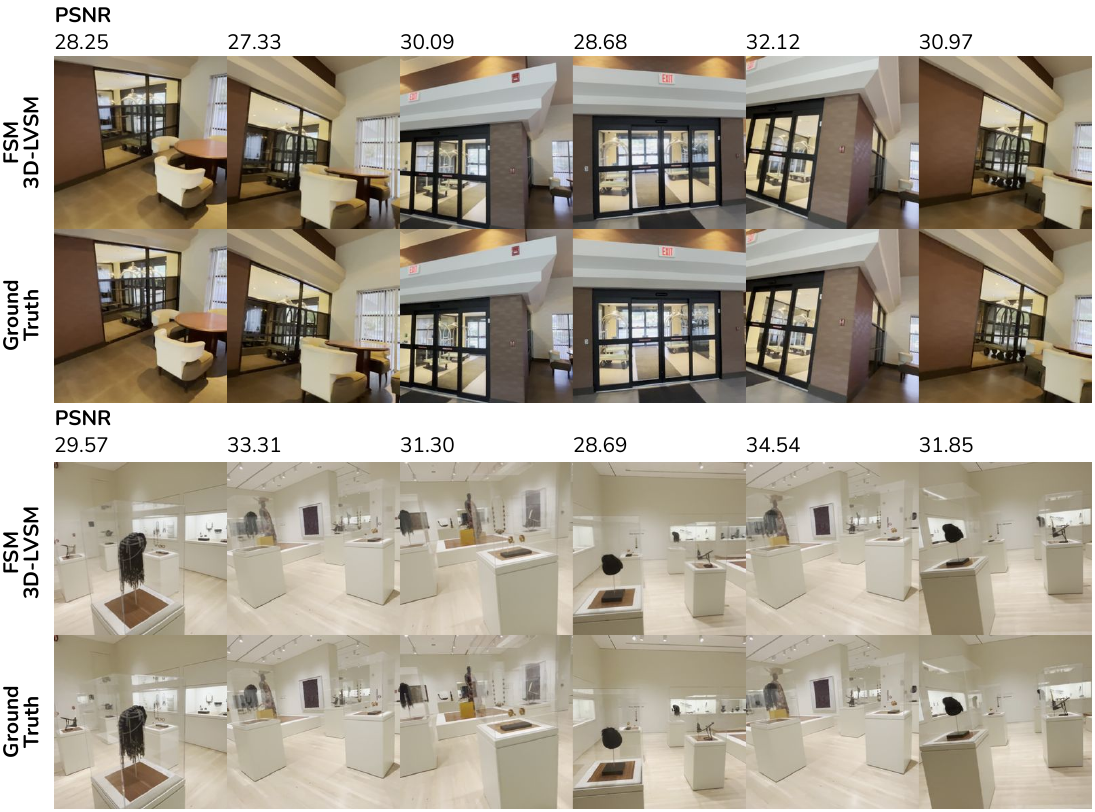}
    \vspace{-15pt}
    \caption{Qualitative results on DL3DV-140 benchmark. \vspace{-5pt}}
    \label{fig:dl3dv_app}
\end{figure*}